\documentclass{article}

\usepackage[final,nonatbib]{neurips_2020}

\usepackage[utf8]{inputenc} 
\usepackage[T1]{fontenc}    
\usepackage{hyperref}       
\usepackage{url}            
\usepackage{booktabs}       
\usepackage{amsfonts}       
\usepackage{nicefrac}       
\usepackage{microtype}      

\usepackage{mathtools}  
\usepackage{amsmath}
\usepackage{amssymb}
\usepackage{tabulary}
\usepackage{multirow}
\usepackage{booktabs}
\usepackage{subcaption}

\usepackage{wrapfig}
\usepackage{etoolbox}

\usepackage{xcolor}
\usepackage{appendix}
\usepackage{enumitem}
\usepackage{fdsymbol}
\usepackage[numbers]{natbib}
\usepackage{multicol}

\newcommand\blfootnote[1]{%
  \begingroup
  \renewcommand\thefootnote{}\footnote{#1}%
  \addtocounter{footnote}{-1}%
  \endgroup
}

\hypersetup{
    colorlinks=true,
    citecolor=black,
    linkcolor=black,
    filecolor=magenta,      
    urlcolor=cyan,
}
    
\title{
Modeling Task Effects on Meaning Representation \\ in the Brain via Zero-Shot MEG Prediction
}

\author{%
  Mariya Toneva{\large \thanks{Equal contribution and joint lead authorship.}} \ \ \textsuperscript{1,2}, Otilia Stretcu{\large\footnotemark[1]} \ \ \textsuperscript{1}, Barnabás Póczos\textsuperscript{1}, Leila Wehbe\textsuperscript{1,2}, Tom M. Mitchell\textsuperscript{1,2} \\
  \textsuperscript{1} Machine Learning Department, Carnegie Mellon University, Pittsburgh, USA \\
  \textsuperscript{2} Neuroscience Institute, Carnegie Mellon University, Pittsburgh, USA \\
   \texttt{mktoneva@cs.cmu.edu, ostretcu@cs.cmu.edu}
}

\begin{document}

\maketitle

\begin{abstract}
 How meaning is represented in the brain is still one of the big open questions in neuroscience. Does a word (e.g., \emph{bird}) always have the same representation, or does the task under which the word is processed alter its representation (answering \emph{``can you eat it?''} versus \emph{``can it fly?''})? The brain activity of subjects who read the same word while performing different semantic tasks has been shown to differ across tasks. However, it is still not understood how the task itself contributes to this difference. In the current work, we study Magnetoencephalography (MEG) brain recordings of participants tasked with answering questions about concrete nouns. We investigate the effect of the task (i.e. the question being asked) on the processing of the concrete noun by predicting the millisecond-resolution MEG recordings as a function of both the semantics of the noun and the task. Using this approach, we test several hypotheses about the task-stimulus interactions by comparing the zero-shot predictions made by these hypotheses for novel tasks and nouns not seen during training. We find that incorporating the task semantics significantly improves the prediction of MEG recordings, across participants. The improvement occurs $475-550$ms after the participants first see the word, which corresponds to what is considered to be the ending time of semantic processing for a word. These results suggest that only the end of semantic processing of a word is task-dependent, and pose a challenge for future research to formulate new hypotheses for earlier task effects as a function of the task and stimuli.
\end{abstract}

\blfootnote{Code available at \href{https://github.com/otiliastr/brain\_task\_effect}{\texttt{https://github.com/otiliastr/brain\_task\_effect}}.}

\section{Introduction}
One of the central goals of artificial intelligence (AI) is to build intelligent systems that understand the meaning of concepts and use it to perform tasks in the real world. Despite the great strides in learning representations, there are still many problems that could benefit from further improvements in understanding and representing \textit{meaning}, such as symbol grounding, common-sense reasoning, and natural language understanding. While machines are limited in these areas, we do have one system that is capable of representing meaning and performing these tasks well: the human brain. Thus, looking to the brain for insights about how we represent and compose meaning may be beneficial. 

Studies of meaning representation in neuroscience have revealed that the brain accesses meaning differently depending on the demands of a task \citep{binder2009semantic,gan2013effect,hebart2018representational,xu2018doctor,wang2018representational}. For instance, the recorded brain activity of a participant that observes the word \textit{``cat''} differs according to whether the participant is asked to answer whether \textit{``cat''} is an animal or a vegetable \citep{kiefer2001perceptual}. The difference is shown to occur between $400-600$ms after \textit{``cat''} is presented to the participant, a period when it is believed that the brain processes the semantics of the perceived word \citep{helenius1998distinct}, suggesting an interaction between the task and stimulus meaning. One hypothesis for the interaction that has received some experimental backing is that, in order to solve the task, the brain uses attention mechanisms to emphasize task-relevant information \citep{smith1974structure,cohen1990control,kanwisher2000visual,cukur2013attention,nastase2017attention}. However, the computational principle behind this attention mechanism is poorly understood, as it can be due to several neural properties, such as an increased response gain, sharper tuning \citep{brouwer2013categorical}, or a tuning shift \citep{cukur2013attention}.

In this work, we propose the first computational model that implements precise hypotheses for the interaction between the semantics of tasks and that of individual concepts in the brain, and tests their ability to explain brain activity. We posit that formulating such a computational model will be a helpful step towards specifying a full account of the task-stimulus interactions. Specifically, we study how tasks interact with the semantics of concepts by building models that predict recorded brain activity of people tasked with answering questions (e.g., \textit{``is it bigger than a microwave?''}) about concrete nouns (e.g., \textit{``bear''}). 
Importantly, the proposed model is able to generalize to previously unseen tasks and stimuli, allowing us to make zero-shot predictions of brain recordings.

Using this computational framework, we show that models that predict brain recordings as a function of the task semantics significantly outperform ones that do not during time windows ($475-550$ms and $600-650$ms) which largely coincide with the end of semantic processing of a word, typically thought to last until $600$ms \citep{helenius1998distinct}. This result suggests that only the end of semantic processing of a word becomes task-dependent and that this effect is related to the meaning of the task. 
We believe that in addition to this result, neuroscientists will also be interested in the ability to computationally compare different hypotheses for the task-stimuli interactions, and we hope that our general problem formulation will benefit future research attempting to study other forms of interaction not considered in this work. 
Additionally, our work may be of interest to the AI community. Further understanding task effects on concept meaning in the brain may provide insights into building AI models that learn how to combine representations specific to the task with task-invariant representations of concepts, as a step towards composing meaning that is both goal-oriented and more easily adaptable to new tasks.

Our \textbf{main contributions} can be summarized as follows:
\begin{itemize}[leftmargin=*]
    \item We propose a means of representing the semantics of the question task that shows a significant relationship with the elicited brain response. We believe such an approach could be useful to future studies on question-answering in the brain.
    \item We provide the first methodology that can predict brain recordings as a function of \emph{both} the observed stimulus and question task. This is important because it will not only encourage neuroscientists to formulate mechanistic computational hypotheses about the effect of a question on the processing of a stimulus, but also enable neuroscientists to test these different hypotheses against each other by evaluating how well they can align with brain recordings. While we have implemented and compared several hypotheses for this effect, and have found some to be better than others, parts of the MEG recordings remain to be explained by future hypotheses. We hope neuroscientists will build on our method to formulate and test such future hypotheses. We make our code publicly available to facilitate this.
    \item We perform all learning in a zero-shot setting, in which neither the stimulus nor the question used to evaluate the learned models is seen during training (i.e. not just as the specific stimulus-question pair but also in combination with any other question/stimulus). Note that this is not the case in previous work that examines task effects, and we are the first to demonstrate how zero-shot learning can be applied successfully to this question. This is important for scientific discovery because it can test the generalization of the results beyond the experimental stimuli and tasks.
    \item We show that models that integrate task and stimulus representations have significantly higher prediction performance than models that do not account for the task semantics, and localize the effect of task semantics largely to time-windows in $475-650$ms after the stimulus presentation.
\end{itemize}

\section{Related work}
Classical neuroimaging experiments that study meaning by contrasting different stimulus conditions often include a task that is related to processing the meaning of the word (such as judging the similarity of two stimuli), however these experiments do not use predictive models that systematically relate the stimulus properties to the brain recordings, and do not explicitly investigate the task effect. 

A number of previous studies have used predictive models to examine the relationship between brain recordings and stimulus properties, but have also not explicitly investigated the effect of a task. In many of these studies \citep{mitchell2008,fyshe2013documents,wehbe14,jain2018incorporating,toneva2019interpreting}, the participants performed only one task -- language comprehension -- and, although this complex task can arguably be broken down into simpler tasks, this question was not explicitly investigated by the authors. In contrast, \citet{sudre2012} explicitly tasked participants with answering yes/no questions about objects. Even though the original paradigm of \citet{sudre2012} results in task-dependent brain recordings, the authors average the brain recordings for the same stimulus across tasks and learn predictive models only based on the semantics of the objects. While averaging over repetitions of the same stimulus can boost the signal-to-noise ratio, it likely loses the task-dependent information in the brain recordings. Here we reanalyze the original task-dependent single-repetition data from \citet{sudre2012} to investigate the task-dependent brain recordings using predictive models that include representations of both the object and the question. 

One previous work uses a predictive model to investigate task effects \citep{cukur2013attention}, and is thus closest to ours. In this work, the authors asked participants to attend to one of two object categories 
in natural scene stimuli. The authors then learn two separate models, each of which is trained to predict the fMRI recordings of participants in one of the $2$ tasks as a function of the stimuli representations. They then compare the learned weights of the $2$ models to conclude that each task-specific model puts more emphasis on those stimulus features that are related to the task. In contrast to this work, we integrate both the task and the stimulus representations into a single zero-shot learning framework, which allows us to predict brain recordings corresponding to novel tasks and stimuli. Additionally, we predict MEG recordings which have a $2000$-times finer temporal resolution than the fMRI recordings used by \citet{cukur2013attention}, which allows us to localize the task effect in time.

The work of \citet{nastase2017attention} also use a computational approach to investigate task effects. These authors account for the task directly in their computational model by constructing a different representational dissimilarity space \citep{kriegeskorte2008representational} for each of two tasks, and then comparing these to the representational dissimilarity space of brain recordings. The representational spaces of the stimuli are entirely task-dependent and do not incorporate the stimulus semantics. This is a limitation because this model is not able to investigate the relationship between the brain recordings and a possible interaction between the task and stimulus. Moreover, similarly to \citet{cukur2013attention}, \citet{nastase2017attention} predict fMRI recordings, limiting the investigation of the task effect in time.

\section{Methodology}
\vspace{-1ex}
\subsection{Brain data}
\begin{figure}
\centering
\begin{minipage}{0.49\linewidth}
  \centering
  \includegraphics[width=\textwidth]{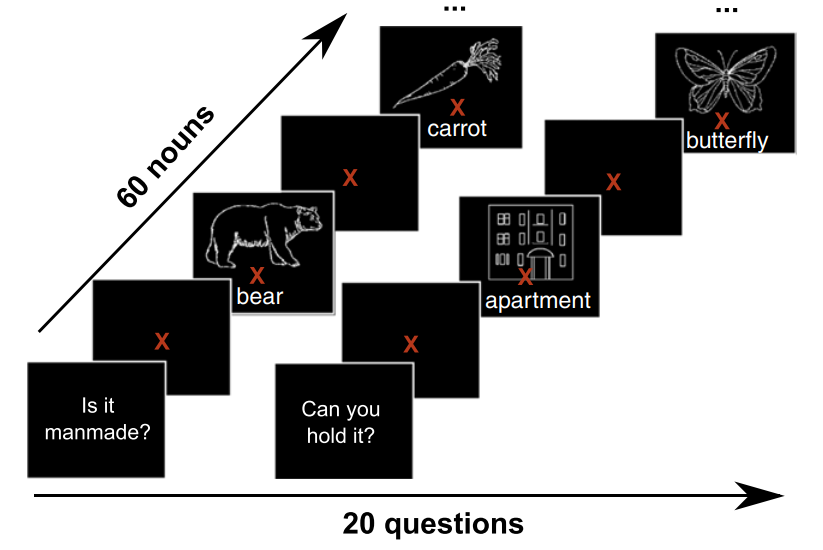}
  \captionof{figure}{Experimental paradigm recreated from \citet{sudre2012}.  Subjects shown a question, followed by $60$ concrete nouns along with their line drawings in random order.
  }
  \label{fig:experiment}
\end{minipage}
\hfill
\begin{minipage}{0.45\linewidth}
  \centering
  \includegraphics[width=\textwidth]{matrix-v3.pdf}
  \captionof{figure}{Feature representations of questions and stimuli obtained from Mechanical Turk.}
  \label{fig:mturk_features}
\end{minipage}
\end{figure}

We aim to study the effect of a task on the brain representation of a stimulus, when the stimulus is shown while performing the task. To this end, investigating the magnetoencephalograpy (MEG) dataset presented in \citet{sudre2012}, which contains $20$ different question tasks, makes for an excellent case study and was provided upon our request.

In this experiment, subjects were asked to perform a question-answering task, while their brain activity was recorded using MEG. 
Figure~\ref{fig:experiment} illustrates the experimental paradigm. Subjects were first presented with a question (e.g., \textit{``Is it manmande?''}), followed by $60$ concrete nouns, along with their line drawings, in a random order. Each stimulus was presented until the subject pressed a button to respond \textit{``yes''} or \textit{``no''} to the initial question. Once all $60$ stimuli are presented, a new question is shown for a total of $20$ questions. 
Thus we have a total of $60\text{ stimuli }\times 20 \text{ questions }= 1200$~examples.

MEG samples the amplitude of the magnetic field induced by neuronal firing at $306$ sensors positioned on the scalp of a subject every millisecond. The data were preprocessed using standard MEG preprocessing procedures (details in Appendix~\ref{appendix:preprocessing}). We analyze the data from the beginning of the stimulus word presentation (i.e. $0$ms) to $800$ms, to avoid contributions to the brain signal from the participant's button-press (median response time across stimuli is $913$ms, averaged across subjects). We further downsample the recordings in time by averaging non-overlapping $25$ms windows, resulting in data of size $306 \text{ sensors } \times 32 \text{ time windows }$. We analyze data from $6$ of the original $9$ subjects. Data from $3$ subjects were excluded because of missing trials.

\subsection{Selecting representations for questions and stimuli}
\label{sec:stim-task-representation}
To study the effect of the question on the meaning representation of a word, we first need a way to represent both the semantics of the question and the word. 
We created two types of word and question representations: one type derived from a pretrained bidirectional model of stacked transformers (BERT) \citep{devlin2018bert}, which is a popular model used for question-answering tasks, and a second type derived from Amazon Mechanical Turk (MTurk) of people answering questions about concrete nouns. We find that the MTurk representations significantly outperform the BERT ones in the prediction tasks outlined in the following sections, and so we focus on the MTurk representations in the main text and provide a detailed description of the BERT features and related results in Appendix \ref{appendix:BERT}. One possible explanation for why representations from BERT perform worse is that BERT may lack commonsense knowledge related to perceptual and visual properties of objects that is necessary to answer the questions in our experiment (e.g. \textit{``Is it bigger than a car?''}). In fact, prior work has shown that BERT representations are deficient of object attributes that are related to questions similar to ours \cite{da2019understanding} and of other physical commonsense knowledge \cite{forbes2019neural}.

The Mechanical Turk data was originally collected by \citet{sudre2012} and was provided at our request. Participants on MTurk were shown a set of $1000$ words (e.g., \textit{``bear'', ``house''}) and were requested to answer $218$ questions about them (e.g., \textit{``Is it fragile?'', ``Can it be washed?''}) on a scale from $1$ to $5$ (``definitely not'' to ``definitely yes''). In this dataset, $60$ out of the $1000$ presented words and $20$ out the $218$ questions corresponded to the stimuli and questions shown during the brain recording experiment.
A complete list of words and questions is shown in Appendix~\ref{appendix:stimuli}.

Using this dataset, we define the representation of a word as a vector containing the MTurk responses for that word to all $198$ questions not in the experiment (see Figure~\ref{fig:mturk_features}). Moreover, we define the task (i.e. question) representation as a vector containing the MTurk responses for $60$ words which are not in the experiment. Using more words did not result in improved performance on the validation set. We purposefully excluded the questions and words in the brain experiment from these representations. Note that \cite{sudre2012} used the same word representations, but to the best of our knowledge, we are the first to represent question semantics as a collection of answers.

With permission from \citet{sudre2012}, we provide the MTurk representations of the stimuli and questions in {\footnotesize \url{https://github.com/otiliastr/brain\_task\_effect}}. We further provide the MTurk human-judgments for all $1000$ words, and the BERT representations discussed in Appendix~\ref{appendix:BERT}.

\subsection{Hypotheses}
\label{subsec:hypotheses}

Next, we formulate several hypotheses of how the question integrates with the stimulus in order to give rise to a task-dependent meaning representation. First, we will introduce the notation used to define the hypotheses, as well as the concrete models described in the next section. Using the notation in Table~\ref{tab:notation}, 
we propose the following hypotheses, also shown in Figure~\ref{fig:models}:

\begin{figure}
    \vspace{-1ex}
    \includegraphics[width=\linewidth]{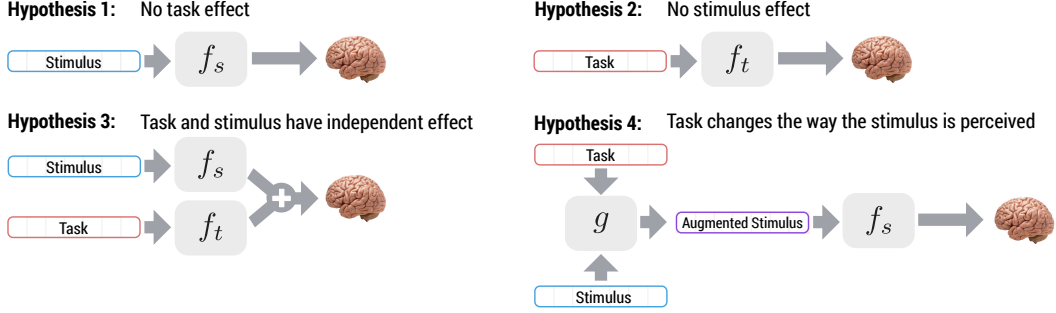}
    \captionof{figure}{Proposed task-stimulus integration hypotheses.}
    \label{fig:models}
    \vspace{-1em}
\end{figure}

\begin{description}[leftmargin=*]
\item \textbf{Hypothesis 1 (no task effect):} The brain activity is not affected by the task and can mostly be explained by the stimulus. Thus, we can approximate the elicited brain activity as: $b = f_s(s)$.

\item \textbf{Hypothesis 2 (no stimulus effect):} The brain activity is not affected by the stimulus and can mostly be explained by the task. Thus, we can approximate the elicited brain activity as: $b = f_t(s)$. While this hypothesis may not predict the brain activity the best, it will allow us to localize the task effect.

\item \textbf{Hypothesis 3 (additive):} Both the stimulus and the task affect the brain activity, but their contributions are independent: $b = f_s(s) + f_t(t)$. 

\item \textbf{Hypothesis 4 (interactive):} The brain activity is well explained by the stimulus, but the task changes the way the stimulus is perceived: $b = f_s(t \otimes s)$. We can think of this as the task focusing \textit{attention} on particular features of the stimulus that are relevant to the task (e.g., in answering the question \textit{``Is it bigger than a car?''} for the stimulus \textit{``dog''}, we pay more attention to the features of \textit{``dog''} that are related to size, and ignore others such as color). This hypothesis aligns with the conclusions of \citet{cukur2013attention} that a task emphasizes those semantic dimensions of the stimulus that are relevant to the task. We use the notation $\otimes$ to represent generically any type of augmentation, and in Section~\ref{sec:predicting-brain} we describe in detail the forms of attention used in our experiments.
\end{description}

\subsection{Predicting brain activity under different hypotheses}
\label{sec:predicting-brain}

We next formulate models to represent the parametric functions $f_s$ and $f_t$ in the proposed hypotheses and to learn the parameters that best predict the brain activity. Our notation is summarized in Table~\ref{tab:notation}. In the rest of this paper, we refer to the hypotheses using the abbreviations H1, H2, H3 and H4.


\subsubsection{Models}
\label{subsubsec:models}

The functions $f_s$ and $f_t$ can be represented using any regression models that map from a feature space to the brain activity space. Prior work \citep{mitchell2008,nishimoto2011reconstructing,wehbe2014, huth2016natural} has shown that simple multivariate regression models such as \textit{ridge regression} are reliable tools for predicting brain activity from stimulus features and are able to achieve good accuracy. For this reason, we will adopt the ridge regression setting for modeling $f_s$ and $f_t$. In ridge regression, we model the output of a function $f$ as a linear combination of the input features: $\hat y = f(x) = xW$, where $W$ is a parameter matrix. $W$ is trained to minimize the loss function $\|Y - XW\|_F^2 + \lambda\|W\|_F^2$, consisting of the mean squared error of the predictions and a regularization term on the parameters to avoid overfitting. Here $X$ and $Y$ represent the training inputs and targets, respectively, stacked together, $\|.\|_F$ denotes the Frobenius norm, and $\lambda > 0$ is a tunable hyperparameter representing the regularization weight. In our setting, the targets of the prediction $Y$ consist of the MEG recording of the brain activity, $Y_b$, described in Table~\ref{tab:notation}. However, the inputs $X$ depend on the hypothesis being tested, as we describe further.

\paragraph{Hypothesis 1:} Under a \textit{no task effect} hypothesis, we predict the brain activity as a function of the stimulus features only, $ Y_b = f_s(X_s) = X_sW_s$, where $W_s \in \mathbb{R}^{F_s \times LT}$. The objective function is:
\begin{align}
    \min_{W_s} \|Y_b - X_sW_s\|_F^2 + \lambda\|W_s\|_F^2
\end{align}

\vspace{-2.5ex}
\paragraph{Hypothesis 2:} Under a \textit{no stimulus effect} hypothesis, we predict the brain activity as a function of the task features only, $ Y_b = f_t(X_t) = X_tW_t$, where $W_t \in \mathbb{R}^{F_t \times LT}$. Our objective function becomes:
\begin{align}
    \min_{W_t} \|Y_b - X_tW_t\|_F^2 + \lambda\|W_t\|_F^2
\end{align}

\vspace{-2.5ex}
\paragraph{Hypothesis 3:} Under an \textit{additive effect} hypothesis, we predict the brain activity as the sum of the stimulus contribution and task contribution: $Y_b = f_s(X_s) + f_t(X_t) = X_s W_s + X_t W_t$. Note that this is equivalent to a single regression function $f(X_s, X_t) = [X_s, X_t] \cdot [W_s; W_t]$, where $[X_s, X_t]  \in \mathbb{R}^{R \times (F_s+F_t)}$ is a concatenation of the stimulus and task features, and $W = [W_s; W_t] \in \mathbb{R}^{(F_s+F_t) \times LT}$ is a concatenation of their corresponding weight matrices. Thus, the objective can be written as: 
\begin{align}
    \min_{W_s,W_t} \|Y_b - [X_s, X_t] \cdot [W_s; W_t]\|_F^2 + \lambda\|[W_s; W_t]\|_F^2
\end{align}

\vspace{-2.5ex}
\paragraph{Hypothesis 4:} Under an \textit{interactive effect} hypothesis, we predict the brain activity as a function of the \textit{augmented} stimulus features. 
The intuition is that the task \textit{augments} the features that are relevant. In this work, we consider an implementation of the augmentation using \textit{soft attention}~\citep{hermann2015teaching}, in which the task reweighs the contribution of the stimulus features. To simplify the notation in the following formulations, we will use $t$ and $s$ to refer to both the identity and the representation of a task and a stimulus in the experiment. Each task $t$ is associated with a set of attention parameters $a_t \in \mathbb{R}^{F_s}$ that rescale the original stimulus features when the stimulus $s$ is presented under question $t$. 
Thus, the augmented stimulus features under question $t$ become $\bar s = a_t \otimes s$, where $\otimes$ represents element-wise multiplication.
The augmented stimuli for all training examples can be stacked together in an a matrix $X_{\bar s}$, and used as input to a ridge regression model, similar to H1:
\vspace{-0.3ex}
\begin{align}
    \min_{W_s} \|Y_b - X_{\bar s}W_s\|_F^2 + \lambda\|W_s\|_F^2
\label{eq:h4_model}
\end{align}
The attention vectors $a_t$ can be precomputed or learned along the regression parameters, as follows:

\begin{table}[t!]
\caption{Notation used in defining the proposed hypotheses and models.}
\label{tab:notation}
\begin{center}
\begin{tabular}{ll|rl}
\toprule[1pt]\midrule[0.3pt]
$N_s$  & num. unique stimuli in experiment, 60                & $\hat b$ & predicted brain activity;  $\hat b \in \mathbb{R}^{LT}$ \\
$N_t$  & num. unique tasks in experiment, 20                  & $X_s$  & stimuli representations, stacked \\
$R$    & total num. repetitions, over all stimuli, 1200       &        & for all repetitions; $X_s \in \mathbb{R}^{R \times F_s}$ \\   
$L$    & space dimension of the brain activity, 306           & $X_t$  & task representations, stacked   \\
$T$    & time dimension of the brain activity, 32             &        & for all repetitions;  $X_t \in \mathbb{R}^{R \times F_t}$    \\
$F_s$  & num. features in stimulus representation,            & $Y_b$  & recorded brain activity, stacked    \\       
       & $198$ for MTurk; $768$ for BERT                      &      & for all repetitions; $Y_b \in \mathbb{R}^{R \times LT}$        \\
$F_t$  & num. features in task representation,                & $f_s$& function mapping from $s$ to $\hat b$; \\
       &       $60$ for MTurk; $768$ for BERT                 &      & $f_s : \mathbb{R}^{F_s} \rightarrow \mathbb{R}^{LT}$  \\
$s$    & stimulus representation; $s \in \mathbb{R}^{F_s}$    & $f_t$& function mapping from $t$ to $\hat b$; \\
$t$     & task representation; $t \in \mathbb{R}^{F_t}$       &      &  $f_t : \mathbb{R}^{F_t} \rightarrow \mathbb{R}^{LT}$     \\
  
\midrule[0.3pt]
\bottomrule[1pt]
\end{tabular}
\end{center}
\vspace{-1em}
\end{table}

\textbf{H4.1. Precomputed attention:} 
The MTurk features have interpretable dimensions for both tasks and stimuli, which enables us to directly compute the hypothesized relevance of different stimuli dimensions to each task. As described in Section \ref{sec:stim-task-representation}, each semantic dimension of a word corresponds to one of the $F_s=198$ non-experimental questions (see Figure \ref{fig:mturk_features}). Given this relationship, we compute the attention parameters for every stimulus presented under task $t$ as
$\smash a_t = \texttt{softmax}([a_{t,\Tilde{t_j}}])$ for 
$\smash j \in \{1,\ldots,F_s\}$, 
where $\smash {\Tilde{t} \in \mathbb{R}^{F_t}} $ is a representation of a non-experimental question, and 
$\smash a_{t,\Tilde{t}} = \texttt{cosine\_similarity}(t,\Tilde{t})$.
We observe that this precomputed attention indeed emphasizes semantically-relevant word features. For example, the word features with highest attention for the question \textit{``Is it made of metal?''} are \textit{``Is it silver?''} and \textit{``Is it mechanical?''}. The top $5$ word features with highest attention for each question are provided in Appendix \ref{appendix:attention}.

\textbf{H4.2. Learned attention:}
We learn the attention parameters together with the regression parameters with the objective of predicting the brain recordings as accurately as possible. A direct approach would be to learn a different set of attention parameters $a_t$ for every task $t$. However, since our goal is to be able to make \textit{zero-shot} predictions for tasks and stimuli never seen during training, we instead learn how to map the features of the task to an attention vector. In our experiments we did so by learning an attention matrix $A \in \mathbb{R}^{F_t \times F_s}$, such that $a_t = \sigma(tA)$, where $\sigma(.)$ represents the sigmoid function, applied element-wise. Putting all pieces together, our objective function becomes:
\vspace{-0.3ex}
\begin{align}
    \min_{W_s, A} \|Y_b - \sigma(X_tA)X_sW_s\|_F^2 + \lambda\|W_s\|_F^2 + \lambda_A\|A\|_F^2
\end{align}
\vspace{-1em}
\subsubsection{Training and evaluation} 

We next train and evaluate all models. Our goal is to predict brain recordings for any new task and word (i.e. zero-shot). Thus, we train all models using \textit{leave-k-out} cross-validation, in which we leave out all training examples that correspond to task-stimuli pairs that contain either a task or a word that will be used for testing. We choose the regularization parameters via nested cross-validation. 

We evaluate the predictions from each model by using them in a classification task on the held-out data, in the \textit{leave-k-out} setting. The classification task is whether we are able to match the brain data predictions for two heldout task-stimuli pairs to their corresponding true brain data. This task has been previously proposed for settings with low signal-to-noise ratio \citep{mitchell2008}. The classification is repeated for each leave-k-out fold and an average classification accuracy is obtained for each sensor-timepoint. We refer to this accuracy as \textit{2v2 accuracy}. The theoretical chance performance is $0.5$.
A more detailed explanation about this metric can be found in Appendix~\ref{appendix:2v2}. Further details about the train/validation/test splitting, parameter optimization, hyperparameter tuning and preventing overfitting can be found in Appendix~\ref{appendix:training-details}.

Our code with all training and evaluation details is available at {\footnotesize \url{https://github.com/otiliastr/brain\_task\_effect}}.

\section{Results and discussion}
\begin{figure}
\centering
\includegraphics[width=0.8\columnwidth]{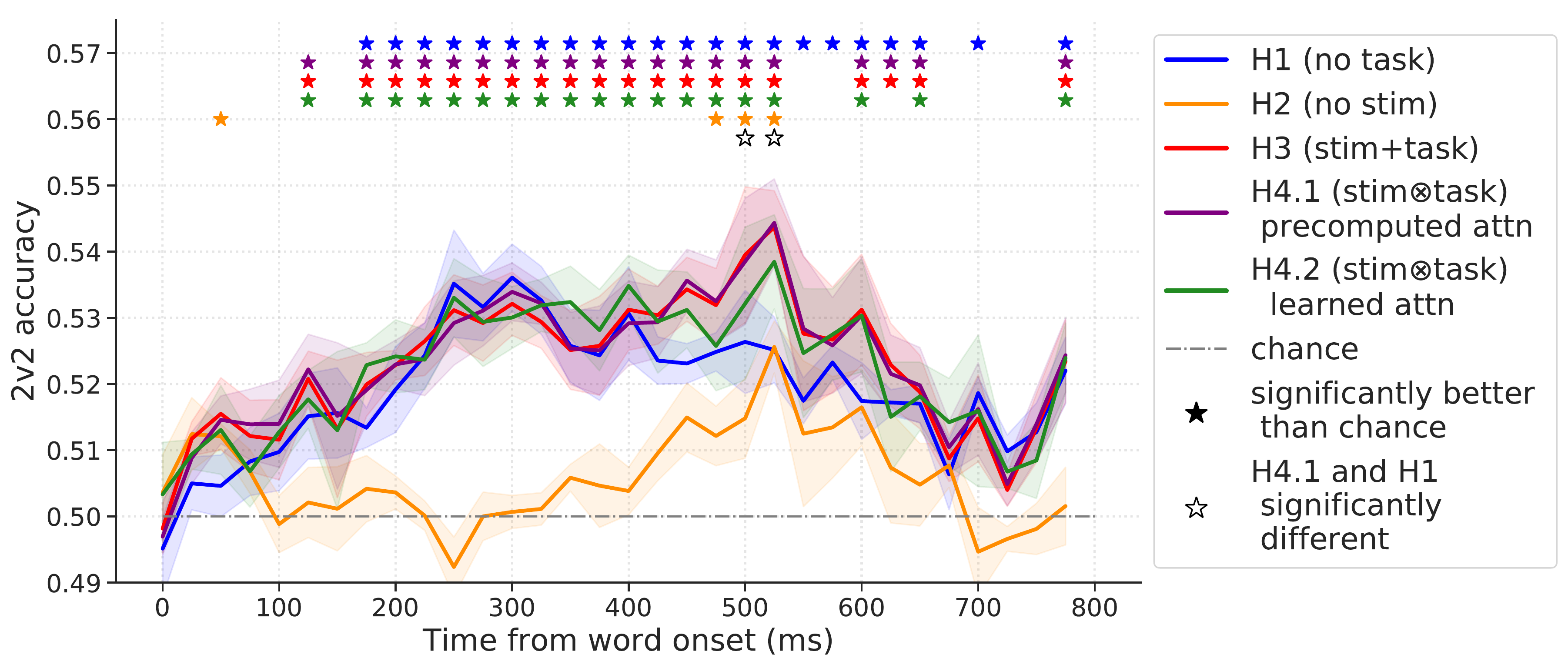}
\captionof{figure}{Performance of all hypotheses at predicting the MEG recordings in $25$ms windows, averaged over sensors. We show the mean and std. error over subjects. The task effect is mostly localized to $475-550$ms. Hypotheses that incorporate both the stimulus and task perform similarly across time.}
\vspace{-1em}
\label{fig:accuracy-per-time}
\end{figure}

\subsection{Effect of incorporating question task semantics}

\textbf{Time window results.} 
We present the 2v2 accuracy per $25$ms time window of all tested hypotheses in Figure \ref{fig:accuracy-per-time}. The time points for which each accuracy significantly differs from chance are indicated with a $\star$ symbol (one-sample t-test, $0.05$ significance level, FDR controlled for multiple comparisons \citep{benjamini1995controlling}). We observe that the hypothesis that only considers the question task semantics (H2) performs significantly better than chance in one early time window ($50-75$ms) and much later during $475-550$ms. The remaining hypotheses also perform better than chance in the same $475-550$ms window, but we observe that during the majority of that time H3 and H4.1 perform significantly better than H1 (paired t-test, 0.05 significance level, FDR controlled for multiple comparisons; significance for the H3-H1 comparison and all other pairwise comparisons are shown in Supplementary Figure \ref{fig:pairwise_hyp} in Appendix \ref{appendix:results}). We conclude that incorporating the question task semantics can improve the prediction of MEG recordings. Note that all discussed times are measured relative to stimulus onset.

\begin{figure}[t!]
\centering
\includegraphics[width=0.43\columnwidth]{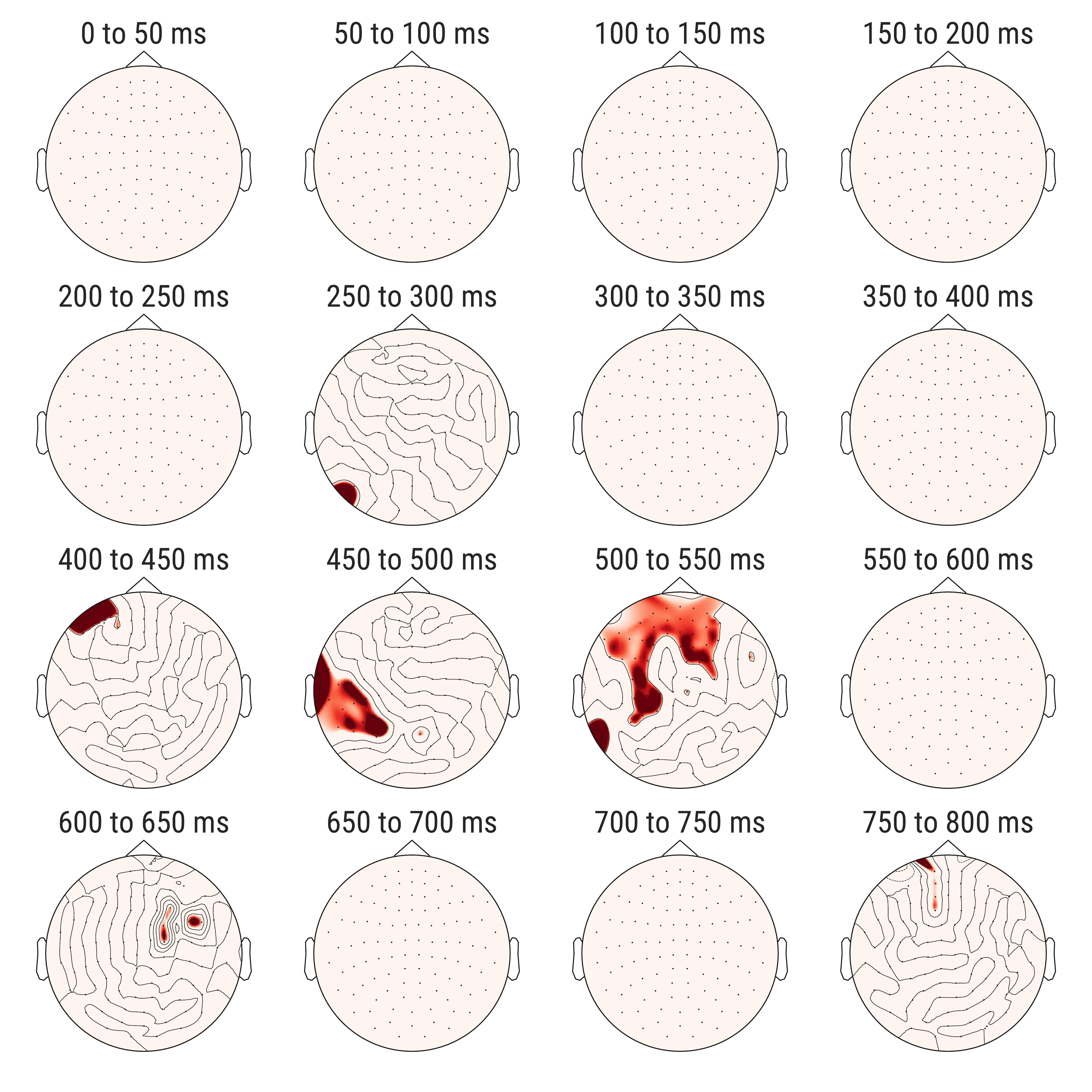}
\includegraphics[width=0.09\columnwidth]{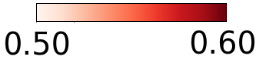}
\includegraphics[width=0.43\columnwidth]{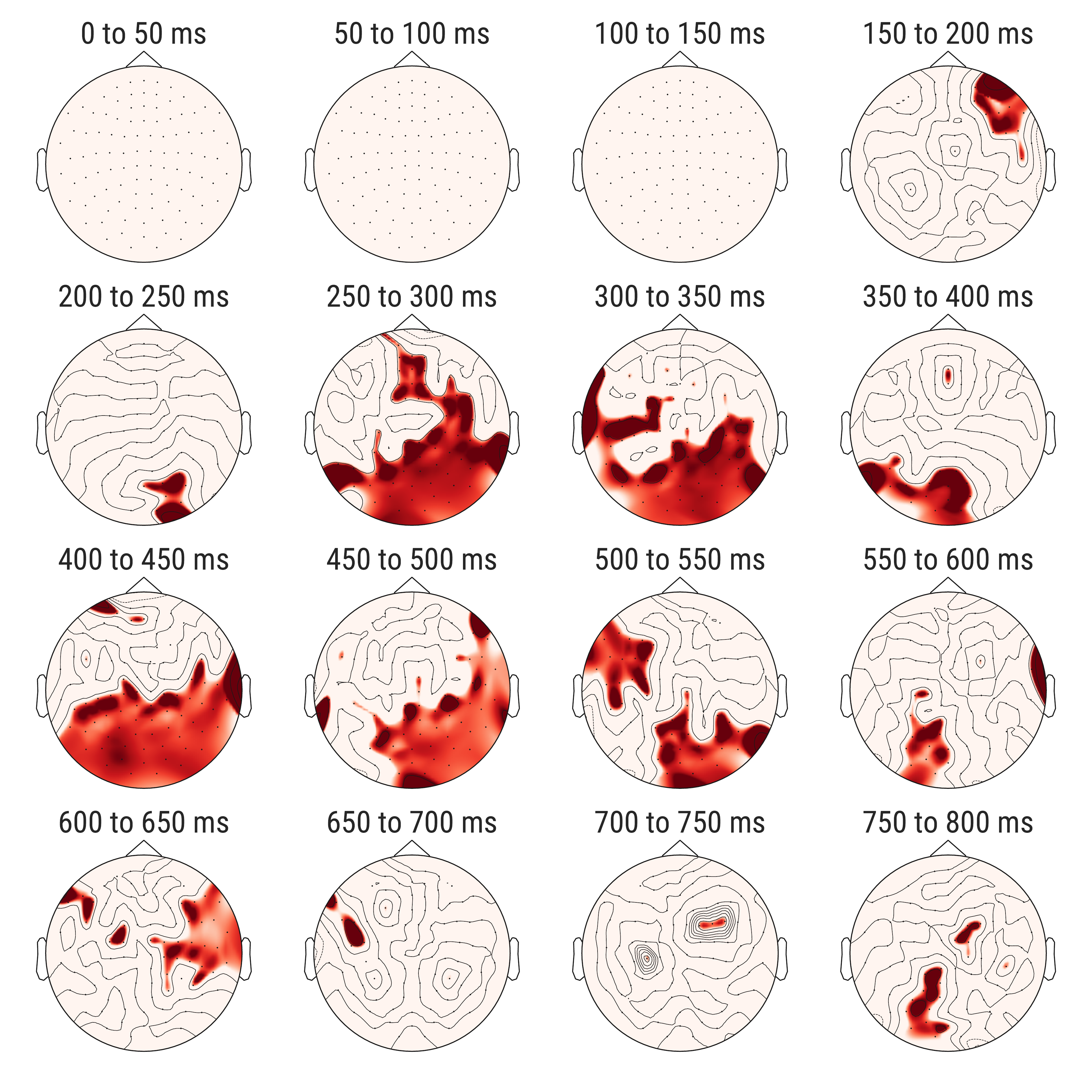}
\caption{Mean 2v2 accuracy across subjects of predicting sensor-timepoints in $50$ms windows using H4.1, when predicting the brain recordings for two word-question pairs that share the same word (Left) and the same question (Right). Displayed accuracies are significantly greater than $0.5$. The main question contribution appears in the frontal and temporal lobes during $400-550$ms, whereas the word contribution is distributed across the occipital and temporal lobes during $200-650$ms.}
\label{fig:h3_brainplot}
\vspace{-1em}
\end{figure}

\textbf{Sensor-timepoint results.} We investigate the task effect further by comparing the contribution of the question-specific precomputed attention and the word features to the accuracy of H4.1 by computing the 2v2 accuracy in two special cases: (1) when the two tested word-question pairs share the same word (i.e. ($q_1$, $w_1$) vs ($q_2$, $w_1$)), higher-than-chance accuracy is attributed to the precomputed attention features; (2) when the two tested word-question pairs share the same question (i.e. ($q_1$, $w_1$) vs ($q_1$, $w_2$)), higher-than-chance accuracy is attributed to the word features. These results are presented per sensor-timepoint in Figure~\ref{fig:h3_brainplot}, where only higher-than-chance accuracies across participants are shown (one-sample t-test, 0.05 significance level, FDR controlled for multiple comparisons). The results are visualized using MNE-Python \cite{gramfort2013meg}. The main contribution of the question-specific attention appears between 400-550ms, localized to the frontal and the left temporal lobes. The contribution of the stimulus features is more distributed, both in time and space. The effect of word semantics begins at $150$ms and extends until the end of the considered time, with major contributions in the occipital lobes ($200-600$ms) and temporal lobes ($400-550$ms, $600-650$ms). For ease of visualization, here we present results for $50$ms time windows. The results for $25$ms time window align with the presented effects and are provided in Appendix \ref{appendix:results}.

\textbf{Beyond MTurk representations.}
We experimented with substituting the Mechanical Turk word and question representations with features extracted from BERT~\citep{devlin2018bert}, which are used by many state-of-the-art methods across several natural language processing problems. In summary, we find that the BERT token-level word embeddings can be a good substitute of the MTurk embeddings, but the question representations of pretrained BERT do not appear to align as closely to the question semantics in the brain. These results can be found in Appendix~\ref{appendix:BERT}. We also tested replacing the stimulus representations with random vectors, which resulted in chance performance.

\subsection{Comparison of task-stimulus interaction hypotheses}
\label{subsec:h4.1_vs_h3}

We further test which of the $3$ hypothesized types of task-stimulus interaction (i.e., independent in H3, precomputed attention in H4.1, or learned attention in H4.2) best explains the observed MEG recordings. We observe that there is no significant difference among these hypotheses when averaging over the performance in all sensors (significance shown in Supplementary Figure \ref{fig:pairwise_hyp}). 

\textbf{Sensor-timepoint results.} A group of sensors in the occipital lobes are significantly better predicted by H3 than by H4.1 at $200-250$ms (paired t-test, $0.05$ significance level, FDR controlled for multiple comparisons). This is when semantic processing of a word begins, so H3 may outperform H4.1 here because H3 has an independent contribution from the word representation. Both H3 and H4.1 perform significantly better than chance during $450-500$ms, and there are different sensor groups in the frontal lobe that are significantly better predicted by each hypothesis than the other. This suggests that this time point may contain both independent and interactive contributions of the task. We lastly observe that H4.1 outperforms H3 in the left temporal lobe during $600-650$ms. This localization suggests that the word and question semantics may interact in this time period, rather than be processed independently. These results are shown in Supplementary Figure \ref{fig:h41_vs_h3}. 

\textbf{Learned attention.} There are no significant differences between H4.2 and H4.1, and the learned attention is highly similar to the precomputed one (Pearson correlation of $0.69$ between the pairwise cosine distances of the task-specific precomputed and learned attention parameters; more details in Appendix \ref{appendix:attention}). This suggests that either the precomputed attention is one optimal way to combine the stimulus and task in predicting the recordings, or that more samples are needed to learn a better one.
To further understand the effect of the sample size, we also evaluated H3 and H4.2 with varying amounts of training data, and found that both models perform increasingly better with more samples. The results and discussion are included in Appendix~\ref{appendix:sample-size-experiments}. 

\subsection{Effect size}
We note that the magnitudes of the presented effects (i.e. accuracies, differences between hypotheses) are limited due to the small amount of data and the underlying difficulty of analyzing single-trial MEG data. The accuracies we observe are on par with other reported single-trial MEG accuracies \cite{wehbe14}. Other work has mitigated the low signal-to-noise ratio of single-trial MEG by averaging the recordings corresponding to different repetitions of the same stimulus \cite{sudre2012} or grouping $20$ examples together for a 20v20 classification task \cite{wehbe14}. Neither is applicable here because our data does not contain repetitions of the same question-stimulus pair, and our zero-shot setting would require us to hold out a large portion of our training set if we were to evaluate on $20$ stimulus-question pairs.

In the absence of these options, we have taken careful precautions to validate our results (by evaluating our models on held-out data in a cross-validated fashion) and evaluated the significance of the model performances and differences between them, and corrected for multiple comparisons. We trust that the effects we have shown to be significant are indeed true, but we note that there may be effects that we are not able to reveal due to limited power and hope that neuroscientists will apply our methods in the future to larger datasets with multiple repetitions.

\subsection{Discussion and relation to previous results}
Taken together, our results point to a robust effect of the question task semantics on the brain activity during $475-550$ms. We also find an effect of the interaction between the question and stimulus semantics during $600-650$ms, localized to the temporal lobe. The temporal lobes are implicated in semantic processing \citep{binder2009semantic, hagoort2013muc, skeide2016ontogeny, hickok2016neural} and specifically in maintaining relevant lexical semantic information for the purposes of integration \citep{hagoort2020meaning}. Since this effect occurs past the time when a word is thought to be processed (i.e. up to $600$ms), it may be related to maintaining specific semantic dimensions that help answer the question (the median response time across participants is $913$ms). In addition to being localized to the temporal lobes, the earlier question effect is also found in the frontal lobes, which are thought to support attention \citep{duncan1995attention, stuss2006frontal}. A task effect that is related to attention is consistent with findings from \citep{cukur2013attention, nastase2017attention}. Our results expand these previous findings by characterizing the temporal dynamics of the task-stimulus interactions.

\section{Conclusions and future work}
We propose a computational framework for comparing different hypotheses about how a task affects the meaning of an observed stimulus. The hypotheses are formulated as prediction problems, where a model is trained to predict the brain recordings of a participant as a function of the task and stimulus representations. We show that incorporating the semantics of a question into the predictive model significantly improves the prediction of MEG recordings of participants answering questions about concrete nouns. The timing of the effect coincides with the end of semantic processing for a word, as well as times when the participant is deciding how to answer the question. 

These results suggest that only the end of semantic processing of a word is task-dependent. This finding may inspire new NLP training algorithms or architectures that keep some computation task-independent, in contrast to current transfer learning approaches for NLP that tune all parameters of a pretrained model when training to perform a specific task \cite{devlin2018bert}.
Moreover, future work can extend our methods to incorporate representations of tasks and stimuli from powerful neural networks that are augmented with improved commonsense knowledge \cite{da2019understanding}, which would eliminate the need for human-judgment annotations. Furthermore, only one of the tested hypotheses (H4.1) is experiment-dependent, while all others can be applied to data from any neuroscience experiment, as long as task and stimulus feature representations can be obtained. Our results pose a challenge for future research to formulate new hypotheses for earlier effects on processing as a function of the task and stimuli.

\section*{Broader Impact}
Our work pursues questions about the function of an average person's brain and makes contributions to basic science. We do not foresee a societal benefit or disadvantage to specific groups of people.

\section*{Acknowledgments}
We thank Gustavo Sudre for collecting the MEG dataset and Dean Pomerleau for collecting the human-judgment Mechanical Turk dataset. This material is based upon work supported by the DARPA D3M program, the NSF Graduate Research Fellowship, the Google Faculty Research Award, start-up funds in the Machine Learning Department at Carnegie Mellon University, and the AFOSR through research grants FA95501710218 and FA95502010118.

\bibliographystyle{unsrtnat}
\bibliography{bibliography}

\appendix
\newpage

\section*{Appendix}

\addcontentsline{toc}{section}{Appendices}
\renewcommand{\thesubsection}{\Alph{subsection}}

\section{Data preprocessing}
\label{appendix:preprocessing}

\paragraph{MEG data preprocessing.}
The data were first preprocessed using the Signal Space Separation method (SSS) ~\citep{taulu2006} in order to isolate the signal components that originate inside of the sensor array. This method was followed by its temporal extension (tSSS) to align the measurements of the head position before each block to a common space. The MEG signal was then filtered using a low-pass filter at $150$Hz and notch filters at $60$Hz and $120$Hz to remove contributions from electrical line noise and other very high frequency noise. Next, the Signal Space Projection method was applied to remove eye blinks, residual movement, and other artifacts.

\paragraph{Input and output normalization.} When training our prediction models, both the model inputs and the targets are normalized as follows. For the input features (which could be word features, question features or both, depending on the hypothesis), we z-score each feature $x_i$ along the sample dimension by assigning $x_i \leftarrow \frac{x_i - \text{mean}(x_i)}{\text{std}(x_i)}$, such that each feature $x_i$ has mean $0$ and standard deviation $1$ across the samples. Similarly, we z-score every sensor-timepoint of the outputs across samples.
However, to make sure our evaluation is correct and no information about the test data has been leaked during training, the mean and standard deviation used when z-scoring is calculated only over the training data.

\section{2v2 metric}
\label{appendix:2v2}
\begin{figure}[h!]
  \begin{center}
    \includegraphics[width=0.7\columnwidth]{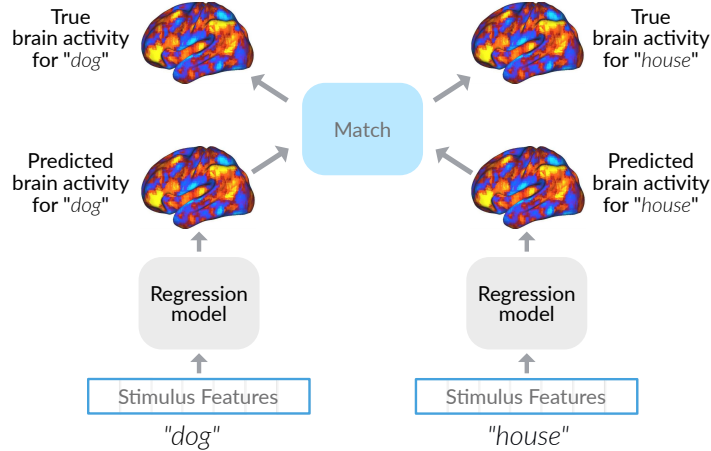}
  \end{center}
  \caption{2v2 metric. The predictions for two repetitions, $\hat{b}_1$ and $\hat{b}_2$, are being matched to their corresponding true brain activities, $b_1$ and $b_2$. The match is performed based on the distances between each of the predictions, to each of the true brain activities: $score_1 = dist(\hat{b}_1, b_1)+dist(\hat{b}_2, b_2)$ and $score_2 = dist(\hat b_1, b_2)+dist(\hat b_2, b_1)$. If $score_1$ < $score_2$, we match $\hat b_1 \leftrightarrow  b_1$ and $\hat b_2 \leftrightarrow b_2$ (correct match, obtaining an accuracy of 1.0), otherwise we match $\hat b_1 \leftrightarrow b_2$ and $\hat b_2 \leftrightarrow b_1$ (wrong match, accuracy 0.0). The distance used in our experiments is cosine distance.}
  \label{fig:2v2}
\end{figure}

There are several metrics that can be used to directly measure how close the brain activity prediction is from the ground truth, such as \textit{Euclidean distance}, \textit{cosine distance}, \textit{percent of variance explained}. In our experiments, we use these metrics to choose the tunable model parameters (e.g., regularization $\lambda$). 
Another way to compare the hypotheses is by trying to match the predicted left-out brain responses to their corresponding  ground truth, as introduced in \citet{mitchell2008}, and illustrated in Figure~\ref{fig:2v2}.
Having 2 left out repetitions, with predictions $\hat b_1$ and $\hat b_2$, and corresponding ground truth $b_1$  and $b_2$, we calculate two scores: 
$score_1 = dist(\hat{b}_1, b_1)+dist(\hat{b}_2, b_2)$ and $score_2 = dist(\hat b_1, b_2)+dist(\hat b_2, b_1)$, where the distance used in our experiments is cosine distance. If $score_1 < score_2$, we match $\hat b_1 \leftrightarrow  b_1$ and $\hat b_2 \leftrightarrow b_2$ (correct match, accuracy 1), otherwise we match $\hat b_1 \leftrightarrow b_2$ and $\hat b_2 \leftrightarrow b_1$ (wrong match, accuracy 0).  We further refer to this as the \textit{2v2 accuracy}. Note that in our setting, we hold out more than 2 repetitions (i.e. word-question combinations), but in order to compute this metric, we take multiple combinations of 2 repetitions.
Under this metric, chance performance is $50\%$.

\section{Training details}
\label{appendix:training-details}

\paragraph{Train/test splits.} The next step is to train and evaluate the proposed models. 
For this, we need to separate our data into a train set and a test set. Since our dataset contains very few samples, as it is usually the case in neuroscience, we adopt the common \textit{leave-k-out} cross-validation approach~\citep{mitchell2008,sudre2012,wehbe2014}, in which the data is repeatedly split into 2 groups: one containing $k$ repetitions for test, and one with $R-k$ repetitions used for training.
Each training set is further split into 2 subgroups using a similar approach: one for training, and one for parameter validation. 
A model is trained on the inner training set using multiple hyperparameters, and the ones with the best average validation accuracy are selected.
Using the best hyperparameters, we retrain on the train+validation data, and compute the final accuracy on the test set.
A common choice for $k$ in approaches that average the stimulus repetitions~\citep{mitchell2008,sudre2012} is $2$, because this allows us to compare the brain activities for two left out stimuli (as described in the next paragraph), while training on as much data as possible. 
However, since we want our models to perform zero-shot learning and to be able to make predictions for both words and questions that have not been seen during training, we leave out from training $2$ stimuli with all their $20$ repetitions under different questions, but also $2$ questions with all $60$ words about which this question was asked (i.e. a total of $2 \times 20 + 60 \times 2 = 160$ examples).
Out of these $160$ examples, we only test the model performance on the word-question pairs for which neither the word nor the question appear in training.
We do this type of splitting both when performing train/test splitting, and for train/validation splitting.

\paragraph{Optimization.} While H1, H2 and H3 can be solved in closed form, we use the Cholesky decomposition approach provided in the Python \texttt{scikit-learn} package \citep{scikit-learn} for computational reasons. In H4, we need to optimize the parameters of the functions $g$ and $f_s$ together, and thus we implemented this using the TensorFlow framework \citep{tensorflow2015-whitepaper} and trained end-to-end using the Adam optimizer \citep{kingma2014adam} with default parameters and learning rate $0.001$. 

\paragraph{Parameters and hyperparameters.} The only hyperparameters in our framework are the regularization parameters $\lambda$ (for all hypotheses) and $\lambda_A$ (only for H4.1). Their values were chosen from the set of values $\{10^{-5}, 10^{-4},..., 10^{7}\}$ using the train/validation/test splitting described above. We also allowed the model to select different $\lambda$ per sensor-timepoint, but found that using the same value for outputs is more stable and leads to better validation accuracy overall. Moreover, we found conducting the parameter validation using the cross-validation setting described above on all subjects and all hypotheses to be prohibitive, and thus we performed the hyperparameter tuning for each hypothesis on a single subject, which was then excluded from testing.
Regarding the number parameters, each hypothesis has a different number of parameters, depending on the size of the of the inputs and extra attention parameters, with H3 > H4.1 > H1 = H4.2 > H2. To ameliorate any effects of overfitting, we allow each hypothesis to choose its own regularization parameters.

\section{BERT features and corresponding results}
\label{appendix:BERT}

\paragraph{Model details} BERT is a bidirectional model of stacked transformers that is trained to predict whether a given sentence follows the current sentence, in addition to predicting a number of input words that have been masked \citep{devlin2018bert}. We use the base pretrained model provided by the Hugging Face Transformers library \citep{Wolf2019HuggingFacesTS}. This model has 12 layers, 12 attention heads, and 768 hidden units.  

\paragraph{Extracting BERT word and question features.} 
We first apply WordPiece tokenization to each of our $60$ word stimuli and questions. 
To extract the word features, we pass the tokens corresponding to each word stimuli into the BERT model and extract the corresponding token-level embeddings. We use these token-level embeddings as the BERT word representations in the following experiments. If any word contains more than $1$ token, it is assigned the average of the corresponding token-level embeddings.

To extract the question features, we pass the tokenization of each question separately into BERT, with a `[CLS]' token and a `[SEP]' token appended to respectively the beginning and end of the tokenized list. This is common practice with inputting multi-word sequences into BERT. We then extracted the hidden layer activations from the CLS token at the last hidden layer, as well as the pooled output. In the pretrained model, the pooled representation of a sequence is a transformed version of the
embedding of the [CLS] token, which is passed through a hidden layer and then a tanh function. 

\paragraph{Question feature dissimilarity.}

\begin{figure}
  \begin{center}
    \includegraphics[width=\columnwidth]{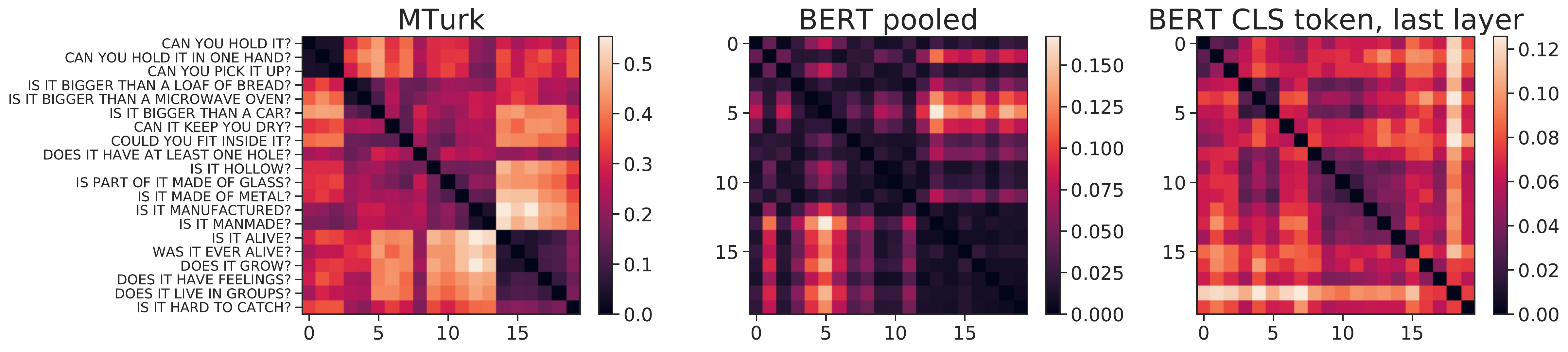}
  \end{center}
  \caption{Pairwise cosine distances among the question representations from MTurk and the two types of BERT question representations. The MTurk representations appear to better cluster semantically-similar questions together.}
  \label{fig:q_dissimilarity}
\end{figure}

There are many different choices for where to extract the question representations from BERT -- e.g. the CLS token representation from any of the $12$ hidden layers, the pooled output, or a mean or max pooling across all token representations in any of the layers. To settle on which BERT representation may be best suited to predict the MEG recordings, we first visualize how similar the representations for our $20$ questions are under different BERT representations (see Figure \ref{fig:q_dissimilarity}). We observe that the MTurk representations appear to better cluster semantically-similar questions together. The representations from the CLS token at the last layer appear to cluster together sentences that share words, whereas the pooled output representations appear to lead to at least two larger clusters that correspond to questions related to animacy and size. We therefore conduct experiments using the BERT pooled output embeddings as the question representations.

\begin{figure}[t]
  \begin{center}
    \includegraphics[width=0.32\columnwidth]{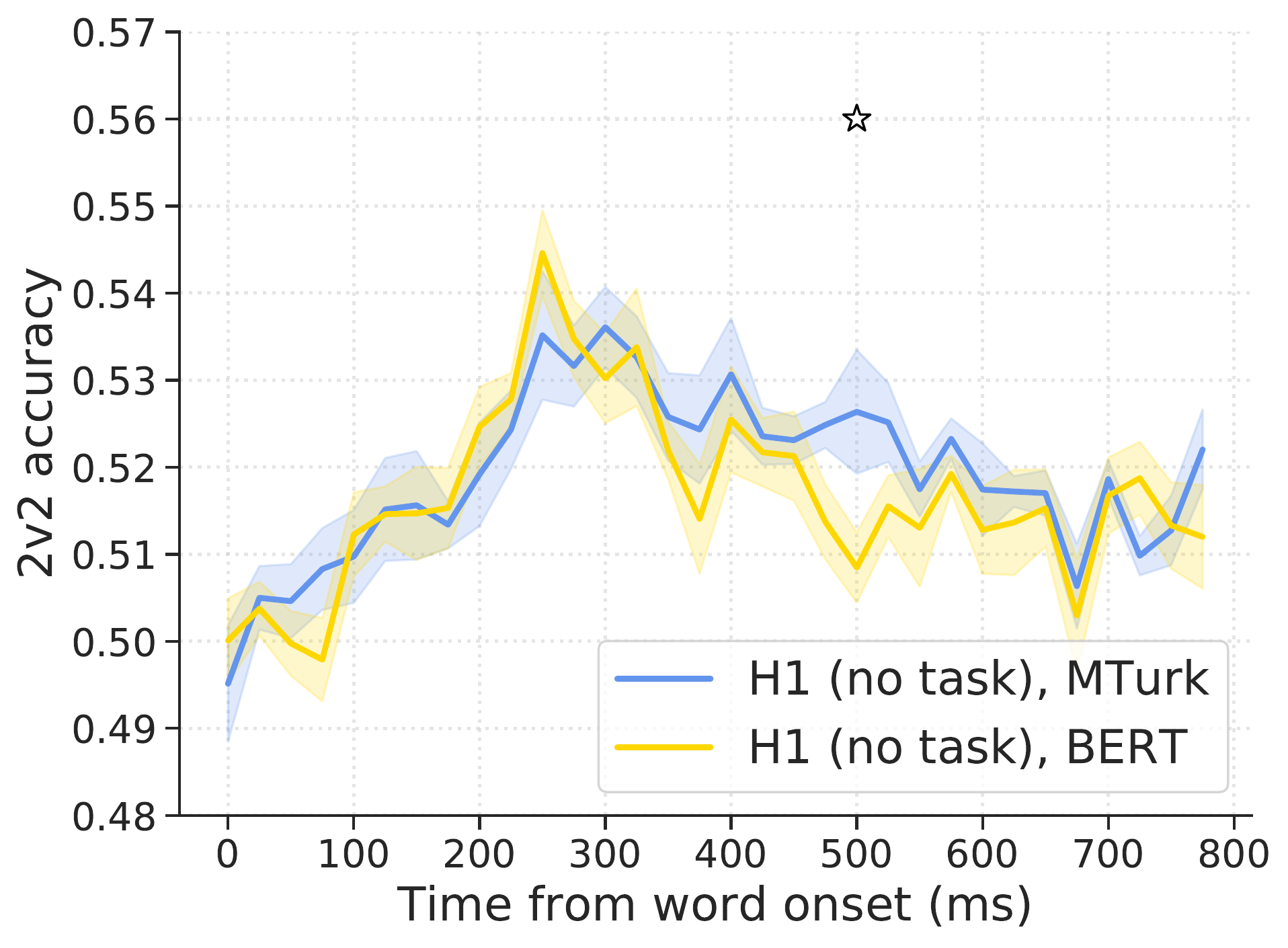}
    \includegraphics[width=0.32\columnwidth]{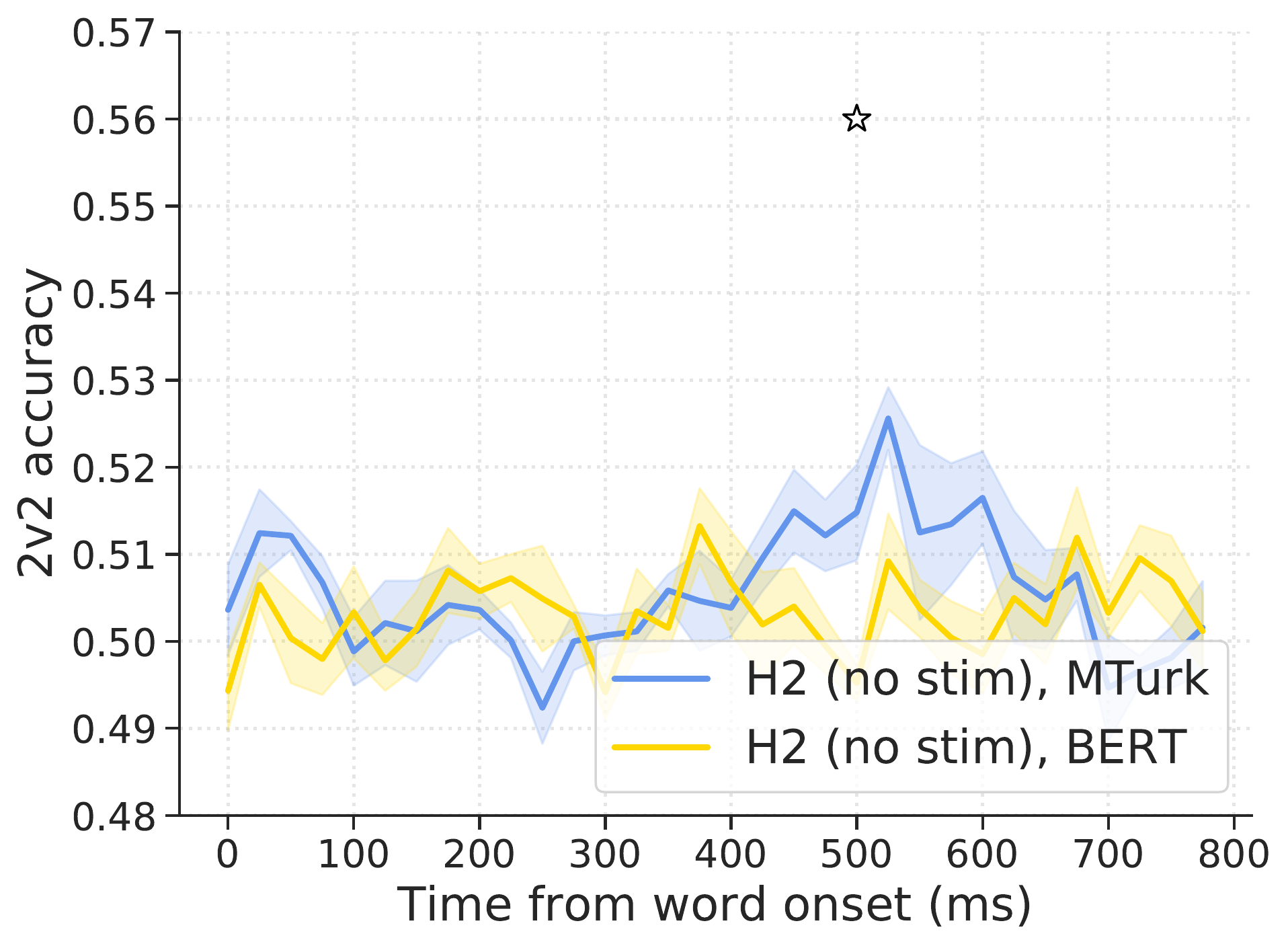}
    \includegraphics[width=0.32\columnwidth]{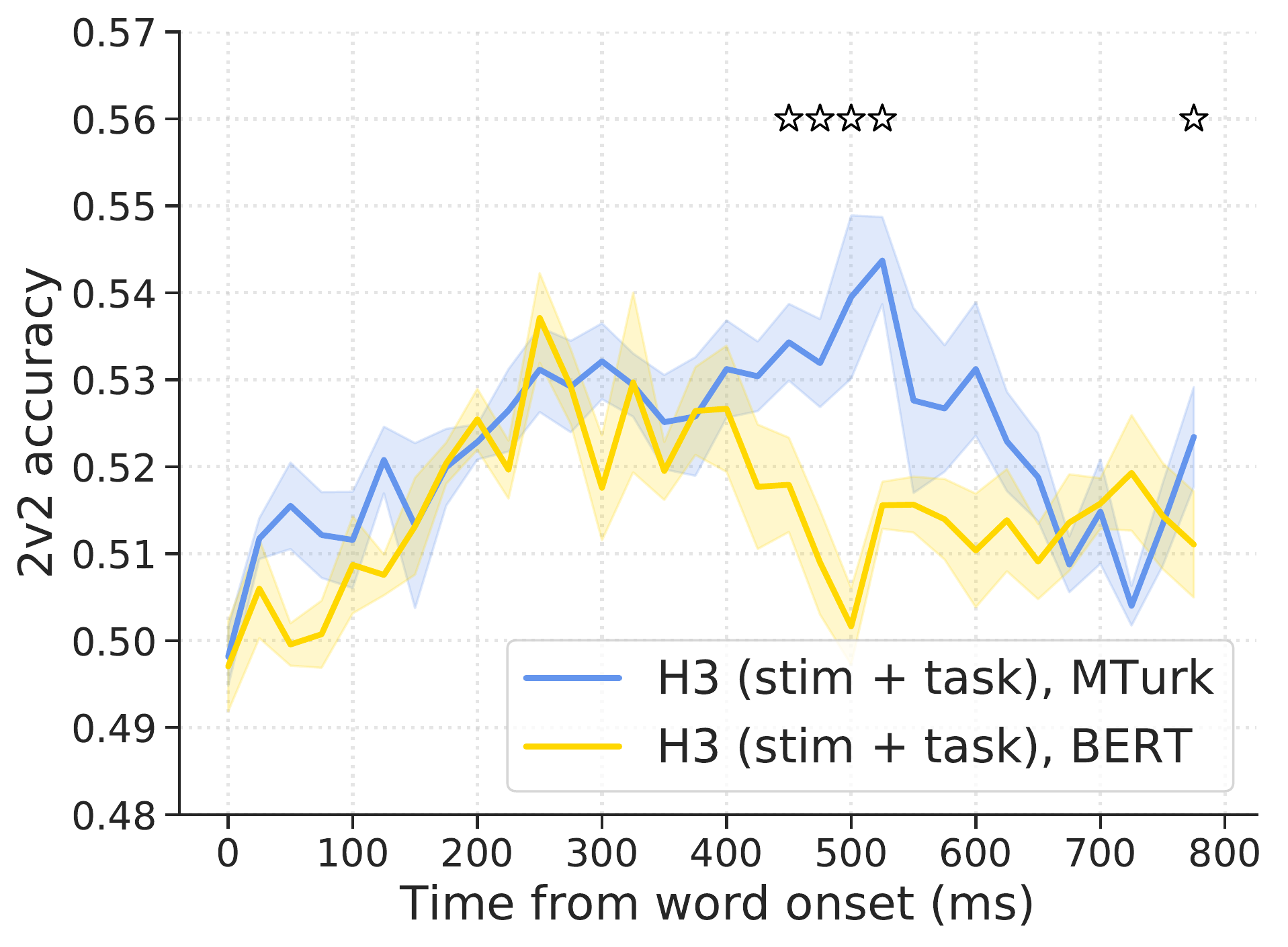}
  \end{center}
  \caption{Comparisons of 2v2 accuracies of predictions computed using MTurk vs. BERT features in each hypothesis. Mean accuracy and standard error across subjects plotted. Points where the means across subjects are significantly different are marked with a $\star$ symbol.}
  \label{fig:mturk_vs_bert}
\end{figure}

\paragraph{Performance against MTurk features.} Features extracted from BERT for both the stimuli and questions perform significantly worse than the Mechanical Turk (MTurk) features in several timewindows across hypotheses (see Figure \ref{fig:mturk_vs_bert}) (paired t-test, significance level 0.05, FDR controlled for multiple comparisons \citep{benjamini1995controlling}). Some of the difference in performance may be due to the large difference in dimensionality between the BERT and MTurk features. The BERT features have much higher dimensionality than the MTurk features, which may lead to more overfitting when using the BERT features. However, this is likely not the only cause for the difference, as the question representations from BERT appear much worse at predicting the MEG recordings in the $450-550$ms timewindow, where we show the question/task semantics contribute most to the MEG recordings. It is likely that the pretrained BERT model is not able to compose the input words in a way that is as brain-aligned as the question representations from Mechanical Turk. It would be an interesting future direction to fine-tune BERT on a question-answering task and compare the performance of the pretrained question representations with that of the fine-tuned BERT representations.

\section{Attention}
\label{appendix:attention}
\paragraph{Relevant word features determined by the precomputed attention.}
The precomputed attention is described in Section \ref{subsubsec:models}, and assigns a different relevance score (between $0$ to $1$) to each word feature for each question. Here we list the top $5$ most relevant word features (i.e. with highest attention scores) for each of the $20$ experimental question, as determined by the precomputed attention. The features are listed in decreasing order of importance (the first is the most important).

    \begin{enumerate}
    \item 'Can you hold it?':
    \begin{itemize}
    \item CAN IT BE EASILY MOVED?
    \item IS IT LIGHTWEIGHT?
    \item WOULD YOU FIND IT IN A HOUSE?
    \item CAN YOU TOUCH IT?
    \item CAN YOU BUY IT?
    \end{itemize}

    \item 'Can you hold it in one hand?':
    \begin{itemize}
    \item CAN IT BE EASILY MOVED?
    \item IS IT LIGHTWEIGHT?
    \item WOULD YOU FIND IT IN A HOUSE?
    \item DO YOU HOLD IT TO USE IT?
    \item CAN YOU BUY IT?
    \end{itemize}
    
    \item  'Can you pick it up?':
    \begin{itemize}
    \item CAN IT BE EASILY MOVED?
    \item IS IT LIGHTWEIGHT?
    \item CAN YOU BUY IT?
    \item WOULD YOU FIND IT IN A HOUSE?
    \item DO YOU HOLD IT TO USE IT?
    \end{itemize}
    
    \item 'Is it bigger than a loaf of bread?':
    \begin{itemize}
    \item IS IT HEAVY?
    \item IS IT TALLER THAN A PERSON?
    \item IS IT LONG?
    \item DOES IT COME IN DIFFERENT SIZES?
    \item IS IT USUALLY OUTSIDE?
    \end{itemize}
    
    \item  'Is it bigger than a microwave oven?':
    \begin{itemize}
    \item IS IT TALLER THAN A PERSON?
    \item IS IT HEAVY?
    \item IS IT BIGGER THAN A BED?
    \item IS IT LONG?
    \item IS IT USUALLY OUTSIDE?
    \end{itemize}
    
    \item  'Is it bigger than a car?':
    \begin{itemize}
    \item IS IT BIGGER THAN A BED?
    \item IS IT BIGGER THAN A HOUSE?
    \item IS IT TALLER THAN A PERSON?
    \item IS IT HEAVY?
    \item IS IT LONG?
    \end{itemize}

    \item  'Can it keep you dry?':
    \begin{itemize}
    \item DOES IT PROVIDE SHADE?
    \item IS IT A BUILDING?
    \item DOES IT PROVIDE PROTECTION?
    \item CAN YOU TOUCH IT?
    \item ARE THERE MANY VARIETIES OF IT?
    \end{itemize}
    
    \item  'Could you fit inside it?':
    \begin{itemize}
    \item IS IT BIGGER THAN A BED?
    \item IS IT TALLER THAN A PERSON?
    \item DOES IT PROVIDE SHADE?
    \item IS IT A BUILDING?
    \item IS IT BIGGER THAN A HOUSE?
    \end{itemize}
    
    \item   'Does it have at least one hole?':
    \begin{itemize}
    \item DOES IT HAVE A FRONT AND A BACK?
    \item IS IT SYMMETRICAL?
    \item DOES IT HAVE PARTS?
    \item DOES IT COME IN DIFFERENT SIZES?
    \item DOES IT HAVE INTERNAL STRUCTURE?
    \end{itemize}
    
    \item   'Is it hollow?':
    \begin{itemize}
    \item IS IT A BUILDING?
    \item DOES IT HAVE FLAT / STRAIGHT SIDES?
    \item DOES IT OPEN?
    \item DOES IT COME IN DIFFERENT SIZES?
    \item CAN YOU TOUCH IT?
    \end{itemize}
    
    \item   'Is part of it made of glass?':
    \begin{itemize}
    \item DOES IT HAVE WIRES OR A CORD?
    \item DOES IT USE ELECTRICITY?
    \item IS IT A BUILDING?
    \item DOES IT HAVE FLAT / STRAIGHT SIDES?
    \item DOES IT HAVE WRITING ON IT?
    \end{itemize}
    
    \item    'Is it made of metal?':
    \begin{itemize}
    \item IS IT SILVER?
    \item IS IT MECHANICAL?
    \item WAS IT INVENTED?
    \item IS IT SHINY?
    \item DOES IT HAVE A HARD OUTER SHELL?
    \end{itemize}

    \item   'Is it manufactured?':
    \begin{itemize}
    \item WAS IT INVENTED?
    \item DOES IT HAVE WRITING ON IT?
    \item DOES IT HAVE FLAT / STRAIGHT SIDES?
    \item CAN YOU USE IT?
    \item CAN YOU BUY IT?
    \end{itemize}
    
    \item   'Is it manmade?':
    \begin{itemize}
    \item WAS IT INVENTED?
    \item DOES IT HAVE FLAT / STRAIGHT SIDES?
    \item DOES IT HAVE WRITING ON IT?
    \item CAN YOU USE IT?
    \item ARE THERE MANY VARIETIES OF IT?
    \end{itemize}
             
    \item   'Is it alive?':
    \begin{itemize}
    \item IS IT CONSCIOUS?
    \item IS IT AN ANIMAL?
    \item IS IT WARM BLOODED?
    \item DOES IT HAVE EARS?
    \item IS IT WILD?
    \end{itemize}
    
    \item   'Was it ever alive?':
    \begin{itemize}
    \item IS IT AN ANIMAL?
    \item IS IT WILD?
    \item CAN IT BITE OR STING?
    \item IS IT CURVED?
    \item IS IT CONSCIOUS?
    \end{itemize}
    
    \item   'Does it grow?':
    \begin{itemize}
    \item IS IT WILD?
    \item IS IT CONSCIOUS?
    \item IS IT AN ANIMAL?
    \item CAN IT BITE OR STING?
    \item IS IT WARM BLOODED?
    \end{itemize}
    
    \item   'Does it have feelings?':
    \begin{itemize}
    \item IS IT CONSCIOUS?
    \item DOES IT HAVE EARS?
    \item DOES IT HAVE A BACKBONE?
    \item IS IT WARM BLOODED?
    \item DOES IT HAVE A FACE?
    \end{itemize}
    
    \item   'Does it live in groups?':
    \begin{itemize}
    \item IS IT AN ANIMAL?
    \item CAN IT JUMP?
    \item IS IT A HERBIVORE?
    \item IS IT WILD?
    \item IS IT CONSCIOUS?
    \end{itemize}
    
    \item    'Is it hard to catch?:
    \begin{itemize}
    \item IS IT FAST?
    \item IS IT A PREDATOR?
    \item IS IT AN ANIMAL?
    \item CAN IT JUMP?
    \item IS IT USUALLY OUTSIDE?
    \end{itemize}
\end{enumerate}

\begin{figure}
  \begin{center}
    \includegraphics[width=\columnwidth]{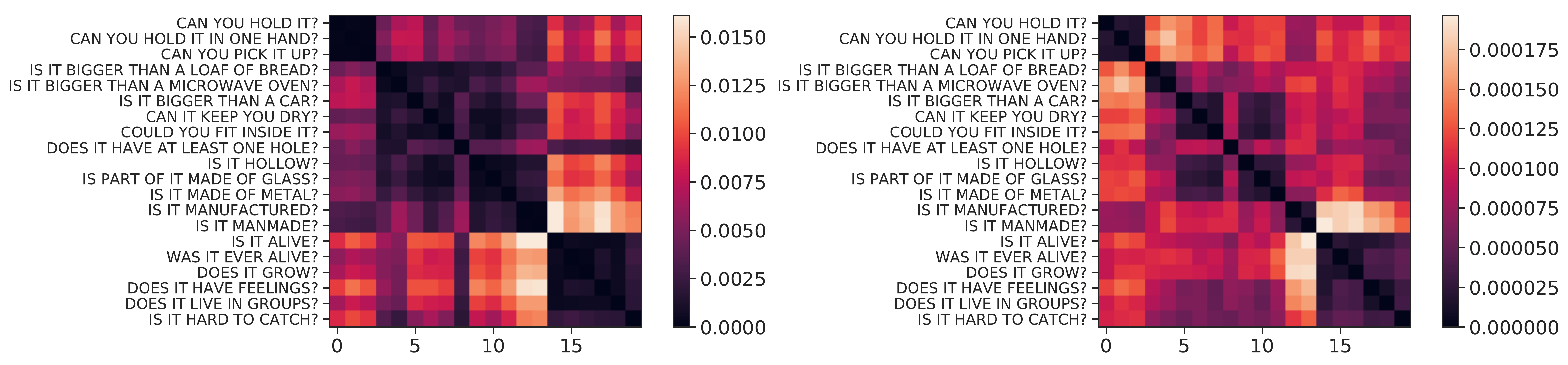}
  \end{center}
  \caption{Pairwise cosine distances across the question-wise attention in H4.1 (Left) and H4.2 (Right). The question-wise attention in H4.2 is an average over participants. The Pearson correlation between these matrices (upper-triangle only) is $0.69$, indicating a high degree of correspondence between the precomputed attention (Left) and the mean learned attention (Right). }
  \label{fig:attn}
\end{figure}

\paragraph{Learned attention vs. precomputed attention} We want to compare the learned attention parameters in H4.2 to the precomputed ones in H4.1. To this end, we compute the pairwise cosine distances across the precomputed question-wise attention, across the learned question-wise attention. The resulting cosine distance matrices are visualized in Figure \ref{fig:attn}. To quantify the similarity between the two, we compute the Pearson correlation between the upper-triangles of the two matrices, which comes out to $0.69$. This indicates that the learned attention, that is randomly initialized, learns to combine the question and word features in a way that is very similar to the precomputed attention.

\section{Supplementary results}
\label{appendix:results}

\begin{figure}
 \centering
 \includegraphics[width=\columnwidth]{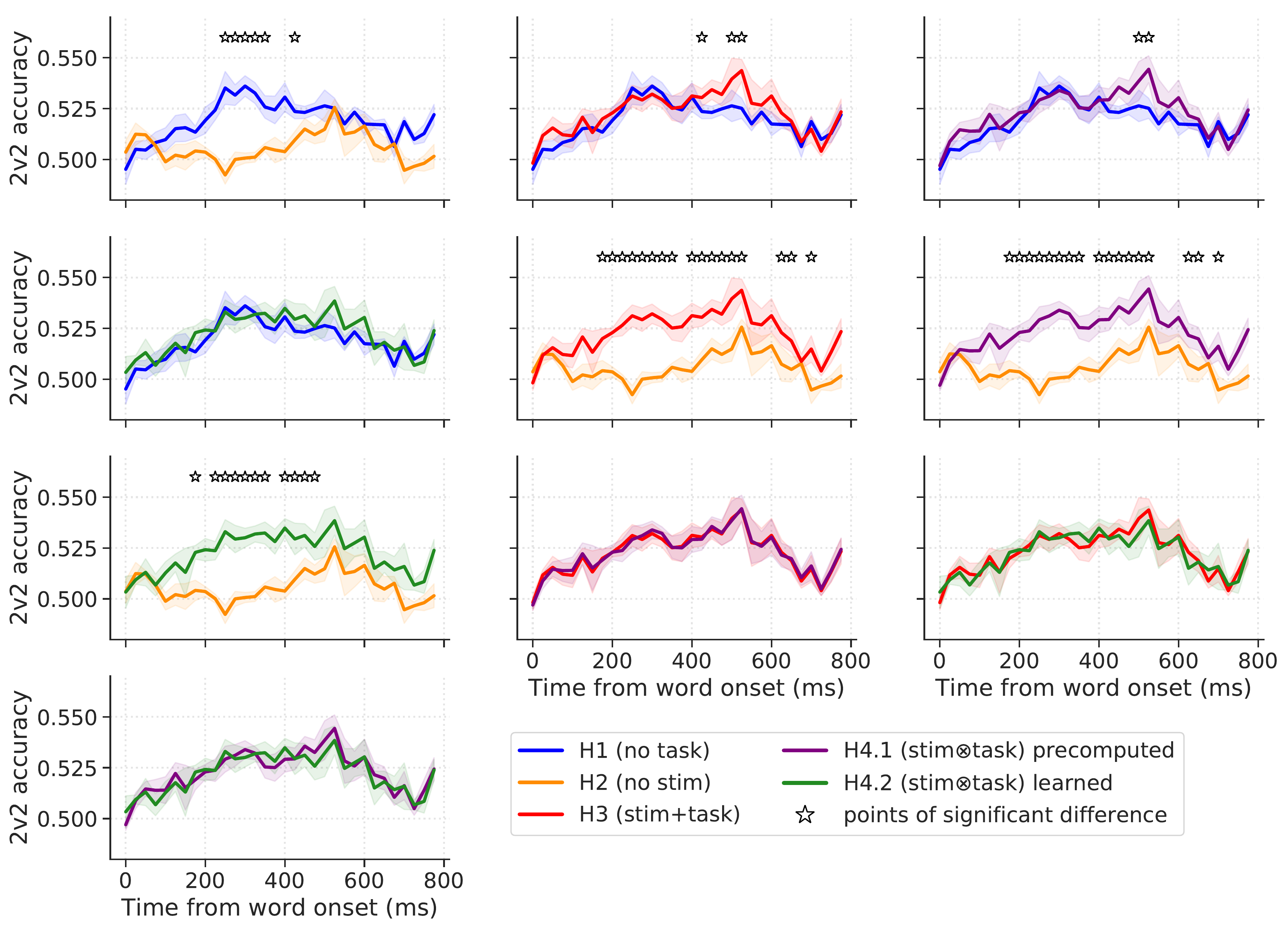}
 \caption{Pairwise comparisons of 2v2 accuracy performance across all tested hypotheses. All timepoints where there is significant difference between the performances of the two displayed hypotheses are marked with a star.}
 \label{fig:pairwise_hyp}
 \end{figure}

\paragraph{Pairwise comparisons across all hypotheses.} We present the pairwise comparisons of 2v2 accuracy performance across all tested hypotheses in Figure \ref{fig:pairwise_hyp}. All timepoints where there is significant difference between the performances of the two displayed hypotheses are marked with a star (paired t-test, significance level 0.05, FDR controlled for multiple comparisons). The hypothesis that does not incorporate information about the stimulus (H2) performs significantly worse than all hypotheses that do during $250-400$ms. The hypothesis that does not incorporate information about the task (H1) performs significantly worse than two hypotheses that do (H3 and H4.1) in time windows between $450$ and $550$ms. There is no significant difference in the performances across timewindows of all hypotheses that are a function of both the stimulus and task (H3, H4.1, and H4.2)

\begin{figure}
\centering
\includegraphics[width=0.43\columnwidth]{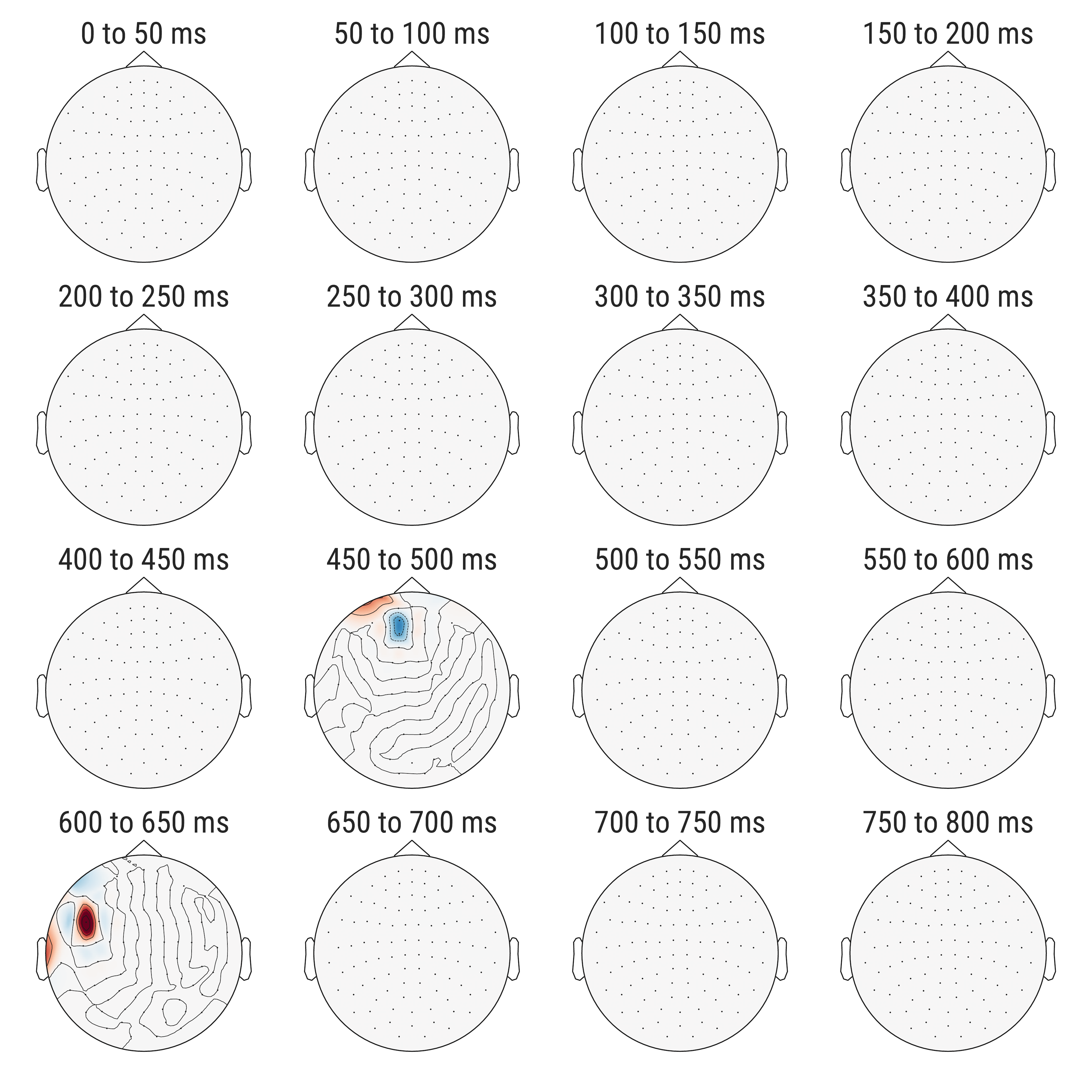}
\includegraphics[width=0.09\columnwidth]{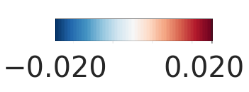}
\includegraphics[width=0.43\columnwidth]{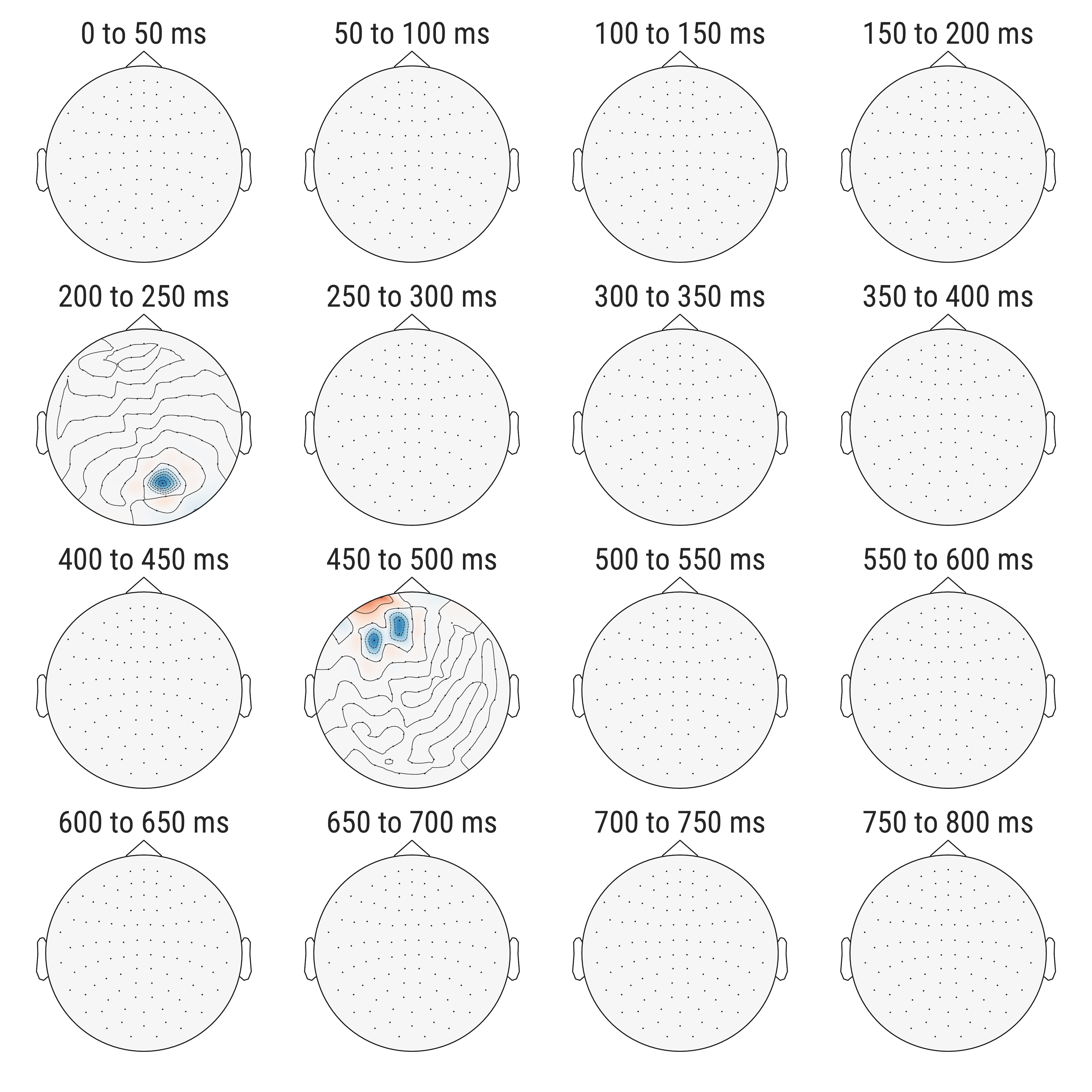}
\caption{Significant differences in performance between H4.1 and H3 per sensor-timepoint (H4.1 accuracy - H3 accuracy). The red points are those where H4.1 significantly outperforms H3, and the blue are those where H3 significantly outperforms H4.1. We're only displaying the significant differences for those timepoints where H4.1 performs significantly better than chance (Left) and where H3 performs significantly better than chance (Right).}
\label{fig:h41_vs_h3}
\end{figure}

\paragraph{Sensor-timepoint comparison of H4.1 against H3.} We present the significant differences in performance between H4.1 and H3 per sensor-timepoint in Figure \ref{fig:h41_vs_h3}. We have plotted H4.1 accuracy - H3 accuracy (the red points are those where H4.1 significantly outperforms H3, and the blue are those where H3 significantly outperforms H4.1). We're only displaying the significant differences for those timepoints where H4.1 performs significantly better than chance (Left) and where H3 performs significantly better than chance (Right). The discussion of these results can be found in the main text in Section \ref{subsec:h4.1_vs_h3}.

\begin{figure}[ht!]
 \centering
 \includegraphics[width=\columnwidth]{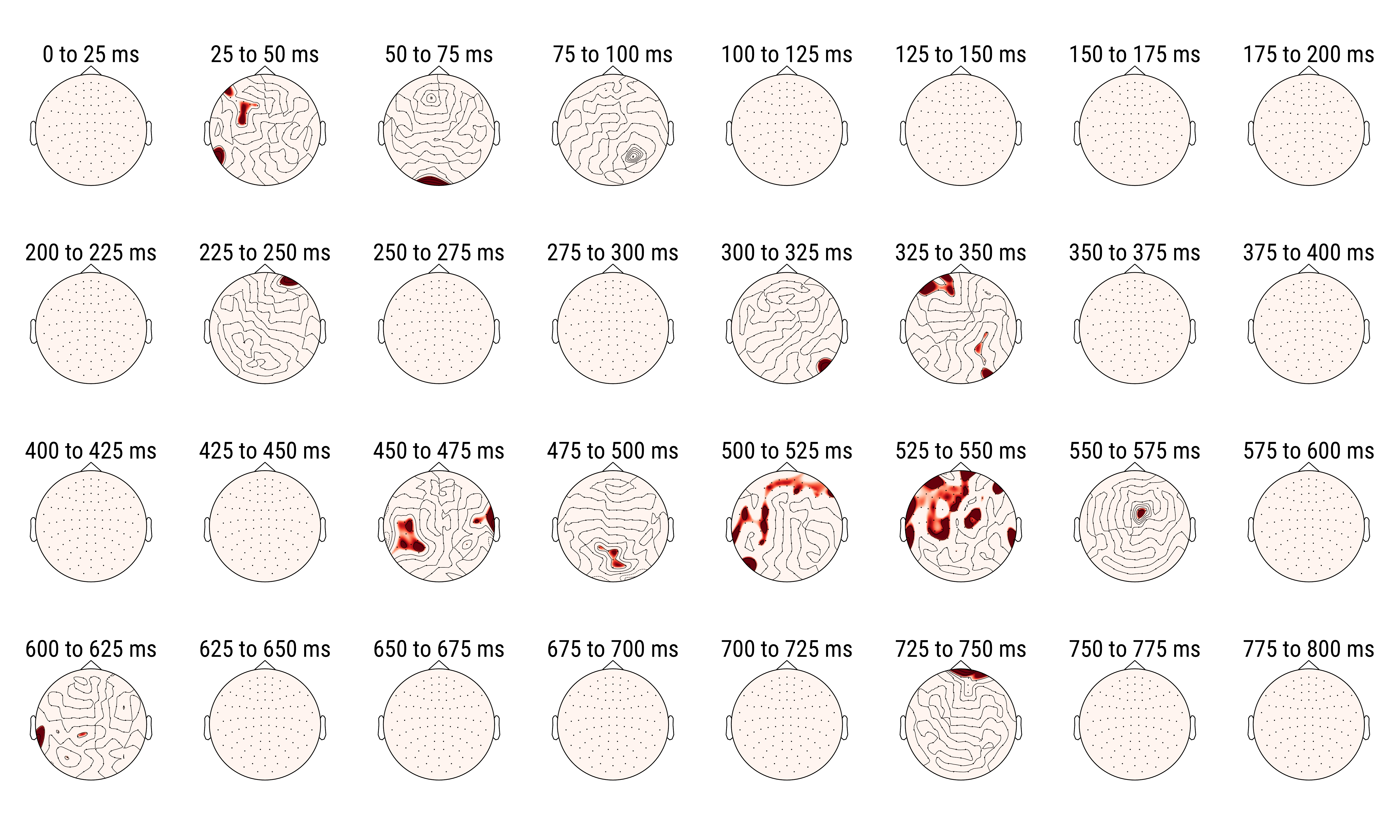}
 \\
  \includegraphics[width=0.2\columnwidth]{colorbar_small.png}
  \\
  \includegraphics[width=\columnwidth]{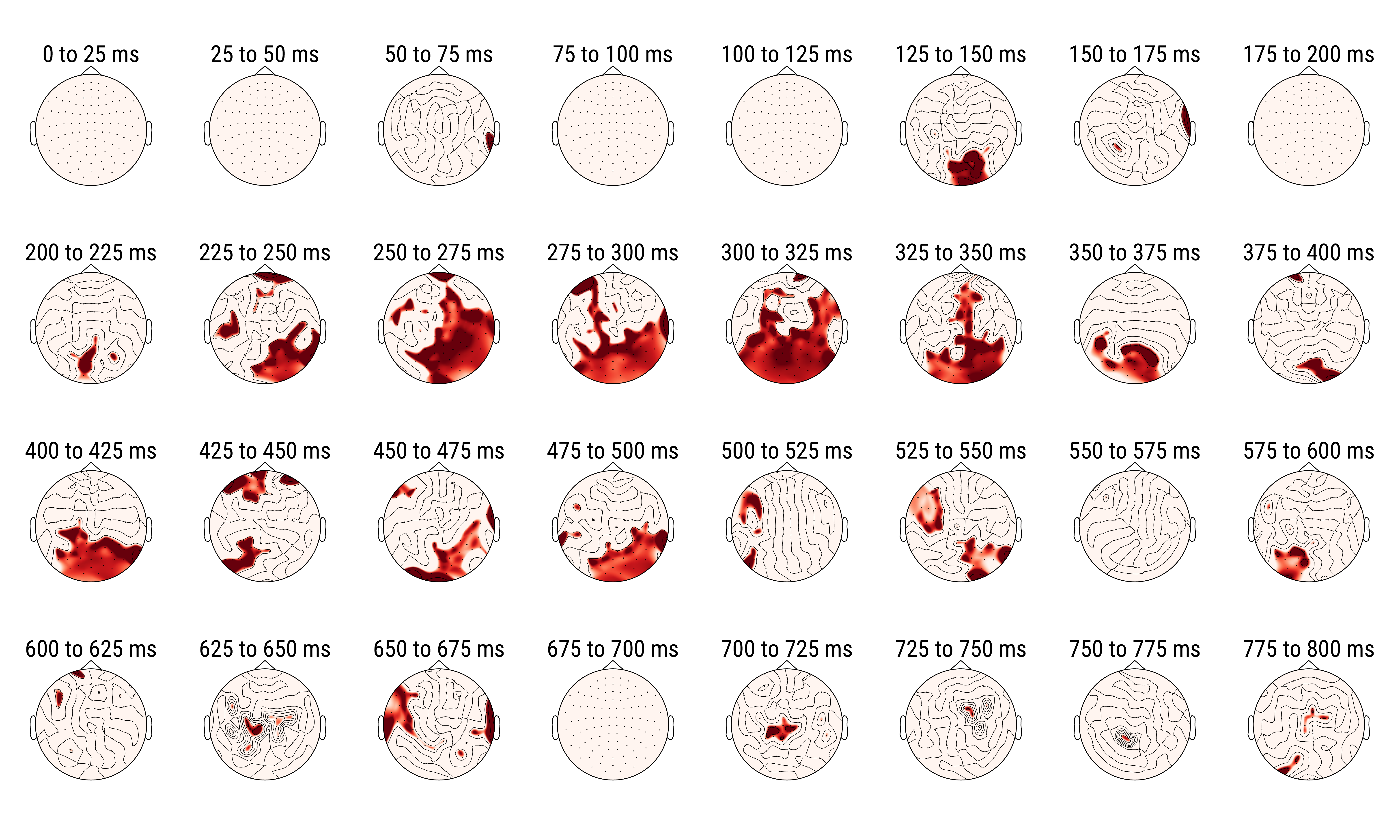}
 \caption{Mean 2v2 accuracy across subjects of predicting sensor-timepoints in 25ms timewindows using H4.1, when predicting the brain recordings for two stimulus-question pairs that share the same word (Top) and the same question (Bottom). The displayed accuracies are significantly greater than chance.}
 \label{fig:25ms_headplot}
 \end{figure}

\paragraph{Sensor-timepoint results for H4.1 for $25$ms windows.} We present the sensor-timepoint results for H3 for $25$ms time-windows in Figure \ref{fig:25ms_headplot}. They follow the general trend of the results from $50$ms time-windows presented in Figure \ref{fig:h3_brainplot}.

\newpage  

\section{Experiments varying the training sample size}
\label{appendix:sample-size-experiments}
We also performed some experiments to compare our different hypotheses using varying amounts of training data.
Since it would be difficult to collect more data beyond the 20 questions and 60 words, we performed this experiment by reducing the amount of data we allow the model to train on. 

\begin{figure}[t]
\centering
\includegraphics[width=\linewidth]{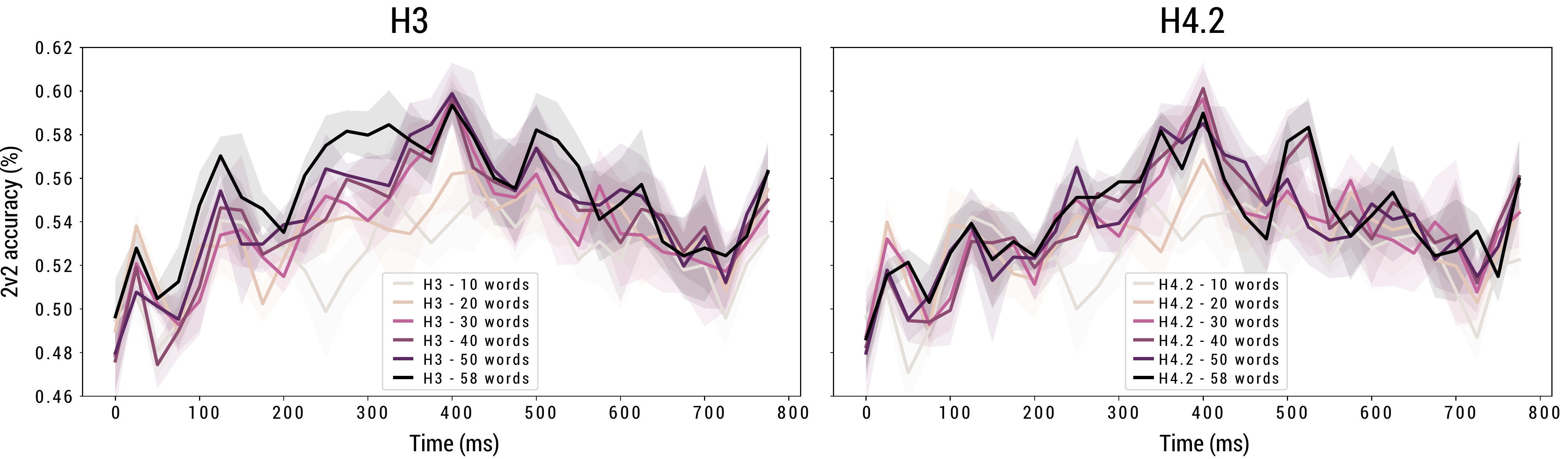}
\caption{Experiments with various amounts of training data.}
\label{fig:vary-training-data-size}
\end{figure}

We trained H4.2 with increasing amounts of data and tested its performance. 
We also tested H3, which is a simpler model that we expect to learn with fewer samples. 
The results are shown in Figure~\ref{fig:vary-training-data-size}.
H3 continues to improve as we add more examples, up to the maximum (i.e. $1044$ samples = $58$words$\times18$ questions).
This suggests that even this simpler model may benefit from more training data.
H4.2 also appears to improve with more samples, however it is less clear whether the performance peak has been reached or whether this is due to the difficulty of the optimization problem.
These results are even more clear in the time interval $450-600$ms, where we expect the two models to perform the best according to the results in~Figure~\ref{fig:accuracy-per-time}.

\section{Stimuli and questions}
\label{appendix:stimuli}
\textbf{Questions in experiment:}

\begin{multicols}{2}
    \begin{enumerate}[noitemsep]
        \item Can you hold it?
        \item Can you hold it in one hand?
        \item Can you pick it up?
        
        \item Is it bigger than a loaf of bread?
        \item  Is it bigger than a microwave oven?
        \item  Is it bigger than a car?
        
        \item  Can it keep you dry?
        \item  Could you fit inside it?
        \item   Does it have at least one hole?
        \item   Is it hollow?
        \item   Is part of it made of glass?
        \item    Is it made of metal?
        
        \item   Is it manufactured?
        \item   Is it manmade?
        
        \item   Is it alive?
        \item   Was it ever alive?
        \item   Does it grow?
        \item   Does it have feelings?
        \item   Does it live in groups?
        \item    Is it hard to catch?
    \end{enumerate}
\end{multicols}

\paragraph{Stimuli in experiment:}
    dog, horse, arm, eye, foot, hand,
   leg, apartment, barn, church, house, igloo, arch,
   chimney, closet, door, window, coat, dress, pants,
   shirt, skirt, bed, chair, desk, dresser, table,
   ant, bee, beetle, butterfly, fly, bottle, cup,
   glass, knife, spoon, bell, key, refrigerator,
   telephone, watch, chisel, hammer, pliers, saw,
   screwdriver, carrot, celery, corn, lettuce, tomato,
   airplane, bicycle, car, train, truck

\textbf{All questions asked on Mechanical Turk:}

\begin{multicols}{3}
    \setlength{\parindent}{2pt}
    \setlength\columnsep{30pt}
    \begingroup
    \obeylines
    \begin{itemize}
        \item  Is it an animal?
        \item  Is it a body part?
        \item  Is it a building?
        \item  Is it a building part?
        \item  Is it clothing?
        \item  Is it furniture?
        \item  Is it an insect?
        \item  Is it a kitchen item?
        \item  Is it manmade?
        \item  Is it a tool?
        \item  Can you eat it?
        \item  Is it a vehicle?
        \item  Is it a person?
        \item  Is it a vegetable / plant?
        \item  Is it a fruit?
        \item  Is it made of metal?
        \item  Is it made of plastic?
        \item  Is part of it made of glass?
        \item  Is it made of wood?
        \item  Is it shiny?
        \item  Can you see through it?
        \item  Is it colorful?
        \item  Does it change color?
        \item  Is one more than one colored?
        \item  Is it always the same color(s)?
        \item  Is it white?
        \item  Is it red?
        \item  Is it orange?
        \item  Is it flesh-colored?
        \item  Is it yellow?
        \item  Is it green?
        \item  Is it blue?
        \item  Is it silver?
        \item  Is it brown?
        \item  Is it black?
        \item  Is it curved?
        \item  Is it straight?
        \item  Is it flat?
        \item  Does it have a front and a back?
        \item  Does it have a flat / straight top?
        \item  Does it have flat / straight sides?
        \item  Is taller than it is wide/long?
        \item  Is it long?
        \item  Is it pointed / sharp?
        \item  Is it tapered?
        \item  Is it round?
        \item  Does it have corners?
        \item  Is it symmetrical?
        \item  Is it hairy?
        \item  Is it fuzzy?
        \item  Is it clear?
        \item  Is it smooth?
        \item  Is it soft?
        \item  Is it heavy?
        \item  Is it lightweight?
        \item  Is it dense?
        \item  Is it slippery?
        \item  Can it change shape?
        \item  Can it bend?
        \item  Can it stretch?
        \item  Can it break?
        \item  Is it fragile?
        \item  Does it have parts?
        \item  Does it have moving parts?
        \item  Does it come in pairs?
        \item  Does it come in a bunch/pack?
        \item  Does it live in groups?
        \item  Is it part of something larger?
        \item  Does it contain something else?
        \item  Does it have internal structure?
        \item  Does it open?
        \item  Is it hollow?
        \item  Does it have a hard inside?
        \item  Does it have a hard outer shell?
        \item  Does it have at least one hole?
        \item  Is it alive?
        \item  Was it ever alive?
        \item  Is it a specific gender?
        \item  Is it manufactured?
        \item  Was it invented?
        \item  Was it around 100 years ago?
        \item  Are there many varieties of it?
        \item  Does it come in different sizes?
        \item  Does it grow?
        \item  Is it smaller than a golfball?
        \item  Is it bigger than a loaf of bread?
        \item  Is it bigger than a microwave oven?
        \item  Is it bigger than a bed?
        \item  Is it bigger than a car?
        \item  Is it bigger than a house?
        \item  Is it taller than a person?
        \item  Does it have a tail?
        \item  Does it have legs?
        \item  Does it have four legs?
        \item  Does it have feet?
        \item  Does it have paws?
        \item  Does it have claws?
        \item  Does it have horns / thorns / spikes?
        \item  Does it have hooves?
        \item  Does it have a face?
        \item  Does it have a backbone?
        \item  Does it have wings?
        \item  Does it have ears?
        \item  Does it have roots?
        \item  Does it have seeds?
        \item  Does it have leaves?
        \item  Does it come from a plant?
        \item  Does it have feathers?
        \item  Does it have some sort of nose?
        \item  Does it have a hard nose/beak?
        \item  Does it contain liquid?
        \item  Does it have wires or a cord?
        \item  Does it have writing on it?
        \item  Does it have wheels?
        \item  Does it make a sound?
        \item  Does it make a nice sound?
        \item  Does it make sound continuously when active?
        \item  Is its job to make sounds?
        \item  Does it roll?
        \item  Can it run?
        \item  Is it fast?
        \item  Can it fly?
        \item  Can it jump?
        \item  Can it float?
        \item  Can it swim?
        \item  Can it dig?
        \item  Can it climb trees?
        \item  Can it cause you pain?
        \item  Can it bite or sting?
        \item  Does it stand on two legs?
        \item  Is it wild?
        \item  Is it a herbivore?
        \item  Is it a predator?
        \item  Is it warm blooded?
        \item  Is it a mammal?
        \item  Is it nocturnal?
        \item  Does it lay eggs?
        \item  Is it conscious?
        \item  Does it have feelings?
        \item  Is it smart?
        \item  Is it mechanical?
        \item  Is it electronic?
        \item  Does it use electricity?
        \item  Can it keep you dry?
        \item  Does it provide protection?
        \item  Does it provide shade?
        \item  Does it cast a shadow?
        \item  Do you see it daily?
        \item  Is it helpful?
        \item  Do you interact with it?
        \item  Can you touch it?
        \item  Would you avoid touching it?
        \item  Can you hold it?
        \item  Can you hold it in one hand?
        \item  Do you hold it to use it?
        \item  Can you play it?
        \item  Can you play with it?
        \item  Can you pet it?
        \item  Can you use it?
        \item  Do you use it daily?
        \item  Can you use it up?
        \item  Do you use it when cooking?
        \item  Is it used to carry things?
        \item  Can you pick it up?
        \item  Can you control it?
        \item  Can you sit on it?
        \item  Can you ride on/in it?
        \item  Is it used for transportation?
        \item  Could you fit inside it?
        \item  Is it used in sports?
        \item  Do you wear it?
        \item  Can it be washed?
        \item  Is it cold?
        \item  Is it cool?
        \item  Is it warm?
        \item  Is it hot?
        \item  Is it unhealthy?
        \item  Is it hard to catch?
        \item  Can you peel it?
        \item  Can you walk on it?
        \item  Can you switch it on and off?
        \item  Can it be easily moved?
        \item  Do you drink from it?
        \item  Does it go in your mouth?
        \item  Is it tasty?
        \item  Is it used during meals?
        \item  Does it have a strong smell?
        \item  Does it smell good?
        \item  Does it smell bad?
        \item  Is it usually inside?
        \item  Is it usually outside?
        \item  Would you find it on a farm?
        \item  Would you find it in a school?
        \item  Would you find it in a zoo?
        \item  Would you find it in an office?
        \item  Would you find it in a restaurant?
        \item  Would you find in the bathroom?
        \item  Would you find it in a house?
        \item  Would you find it near a road?
        \item  Would you find it in a dump/landfill?
        \item  Would you find it in the forest?
        \item  Would you find it in a garden?
        \item  Would you find it in the sky?
        \item  Do you find it in space?
        \item  Does it live above ground?
        \item  Does it get wet?
        \item  Does it live in water?
        \item  Can it live out of water?
        \item  Do you take care of it?
        \item  Does it make you happy?
        \item  Do you love it?
        \item  Would you miss it if it were gone?
        \item  Is it scary?
        \item  Is it dangerous?
        \item  Is it friendly?
        \item  Is it rare?
        \item  Can you buy it?
        \item  Is it valuable?
    \end{itemize}
    \endgroup
    \end{multicols}

\end{document}